\documentclass{article}

\usepackage{microtype}
\usepackage{graphicx}
\usepackage{subfigure}
\usepackage{booktabs} 

\usepackage[accepted]{arxiv}

\usepackage{xurl}

\usepackage{hyperref}

\usepackage{amsmath}
\usepackage{amssymb}
\usepackage{mathtools}
\usepackage{amsthm}

\usepackage{paralist}

\usepackage[capitalize,noabbrev]{cleveref}

\theoremstyle{plain}
\newtheorem{theorem}{Theorem}[section]

\theoremstyle{definition}
\newtheorem{definition}[theorem]{Definition}

\theoremstyle{remark}

\usepackage[textsize=tiny]{todonotes}

\usepackage{multicol}
\usepackage{blkarray}
\usepackage{tabularray}

\arxivtitlerunning{Do Attention Heads Compete or Cooperate?}

\begin{document}

\twocolumn[
\arxivtitle{Do Attention Heads Compete or Cooperate during Counting?}




\begin{arxivauthorlist}
\arxivauthor{P\'{a}l Zs\'{a}mboki}{renyi}
\arxivauthor{\'{A}d\'{a}m Frakn\'{o}i}{elte}
\arxivauthor{M\'{a}t\'{e} Gedeon}{bme}
\arxivauthor{Andr\'{a}s Kornai}{bme,sztaki}
\arxivauthor{Zsolt Zombori}{renyi,elte}
\end{arxivauthorlist}

\arxivaffiliation{renyi}{HUN-REN Alfr\'{e}d R\'{e}nyi Institute of Mathematics, Budapest, Hungary}
\arxivaffiliation{elte}{E{\"o}tv{\"o}s Lor\'{a}nd University, Budapest, Hungary}
\arxivaffiliation{bme}{Budapest University of Technology and Economics, Budapest, Hungary}
\arxivaffiliation{sztaki}{HUN-REN Institute for Computer Science and Control, Budapest, Hungary}

\arxivcorrespondingauthor{P\'{a}l Zs\'{a}mboki}{zsamboki@renyi.hu}
\arxivcorrespondingauthor{\'{A}d\'{a}m Frakn\'{o}i}{fraknoiadam@gmail.com}
\arxivcorrespondingauthor{M\'{a}t\'{e} Gedeon}{gedeonm01@gmail.com}
\arxivcorrespondingauthor{Andr\'{a}s Kornai}{kornai@math.bme.hu}
\arxivcorrespondingauthor{Zsolt Zombori}{zombori@renyi.hu}

\arxivkeywords{Machine Learning, ARXIV}

\vskip 0.3in
]



\printAffiliationsAndNotice{}  

\begin{abstract}
We present an in-depth mechanistic interpretability analysis of training small transformers on an elementary task, counting, which is a crucial deductive step in many algorithms. In particular, we investigate the collaboration/competition among the attention heads: we ask whether the attention heads behave as a pseudo-ensemble, all solving the same subtask, or they perform different subtasks, meaning that they can only solve the original task in conjunction. Our work presents evidence that on the semantics of the counting task, attention heads behave as a pseudo-ensemble, but their outputs need to be aggregated in a non-uniform manner in order to create an encoding that conforms to the syntax. Our source code will be available upon publication.
\end{abstract}

\section{Introduction}
\label{sec:intro}

The transformer architecture, via the multistage training method of first
imbuing the model with generic linguistic knowledge and then finetuning it to
the task at hand, has proven remarkably versatile in creating computer
programs that can perform assignments as diverse as emotional support, idea
brainstorming, mathematical problem solving, or pair programming.

The path to these heights does not look like a steady climb: it features
sudden jumps in the form of emergent features. As humanity relies on language
models in tasks of increasing importance, it is paramount to understand how
this phenomenon comes about and how language models are making their
predictions.

The approach of \emph{mechanistic interpretability} aims to do this by zooming
in on the models and studying the circuits that form the model as a program.
This method was first applied to convolutional neural networks
\cite{cammarata2020thread:}. Since then, transformers are also being actively
studied this way, starting with \cite{elhage2021mathematical}, where they
study in-context learning by attention-only transformers, on very simple
information retrieval tasks.

We aim to study the reasoning capabilities of transformers and we
follow the methodology of \citet{Nanda:2023} by training small
transformer models on simple algorithmic tasks, in order to fully
interpret the trained model. We use a simple counting task, provide a
handcrafted minimal solution and, informed by a careful analysis, we
present a hypothesis of how the trained models solve the task. We also
provide interventions to support our hypothesis.

Our investigation centers around the topic of how the trained model's
prediction is composed of that of its components.  We examine what individual
heads learn, how they perform in isolation and how the output layer of the
model learns to exploit them. We bring in a new research question: in a
transformer, do attention heads behave as a pseudo-ensemble, all solving the
same subtask, or do they perform different subtasks, meaning that they can
solve the task only in conjunction?

We introduce a number of probing
experiments to study this problem. Our results indicate that the attention
heads behave as a pseudo-ensemble: they learn to output representations that
make the counting task solvable by logistic regression. However, we show that
in order to conform to the syntax of the full task, i.e. to end the generated sentence,
the output layer of the model needs to aggregate these representations in a non-uniform way.

In summary, our contributions are as follows:

\begin{itemize}
  \item We introduce the Count01 language, which deals with a remarkably
simple counting task, whether a string has more 1s than 0s in the presence of
noise given by 2s. This can be separated to the two subtasks of outputting the answer token (the main task) and ending the sentence (the syntactic task).
\item We
also provide a minimal solution which, however, the model cannot
reach via learning from random initialisation.
\item We introduce the Separation Accuracy (s-acc) metric, using which we can show that parallel heads in an attention layer all aim to learn representations that make the main task solvable via logistic regression.
\item We notice that the output layer does not treat said representations in a uniform way that would exploit this solvability: the logit contributions of each head are highly biased. We introduce another, ROC AUC-based metric to show that, besides this bias, most heads give meaningful contribution to the next token logits.
\item We show that the raison d'être of these biases is to balance logit contributions that solve the main and the syntactic tasks.

\end{itemize}

\section{Related Work}

\label{sec:related}

In a pioneering work on mechanistic interpretability of transformers 
\cite{elhage2021mathematical} investigate in-context
learning effects of 0, 1 and 2-layer attention-only transformers trained via
next token prediction on a text corpus. They point out that "One layer
attention-only transformers are an ensemble of bigram and “skip-trigram”
(sequences of the form "A… B C") models.", but they don't investigate the
formation of this ensemble. Their focus is on induction heads, formed by
2-layer attention-only transformers.

An important current in mechanistic interpretability is studying the circuits formed by small transformers while training on formal tasks. A prime example is \cite{Nanda:2023}, where they teach 1-layer transformers on modular addition. They show that the composite of the 4-head self-attention block and the first layer of the MLP together with the activation function map to trigonometric functions of the input. On the other hand, they do not study how the attention heads cooperate to this end.

Programs written in the Restricted Access Sequence Processing Language (RASP)
\cite{weiss2021thinking} can be implemented by transformer models. This yields
readily interpretable transformer implementations of counting tasks such as
Double Histogram and Shuffle-Dyck. Moreover, it is possible to restrict
transformer models to learn easy to implement programs
\cite{friedman2023learning}. We are more interested in the programs
unrestricted transformer models learn.

Minimal architectural hyperparameters for a single head 1-layer transformer to
learn the Histogram task are studied in \cite{behrens2024counting}. Depending
on the hyperparameter configurations, they hypothesize two possible families
of programs: relation- and inventory-based counting, for which they give
example implementations. However, it is not proven via mechanistic
interpretability that the models actually learn the hypothesized programs. Our
problem statement differs from theirs significantly in that the amount of
counting we perform is limited only by input string length, with vocabulary
size kept to 8 (including [BOS] and [EOS]).

In \cite{wen2024transformers} it is shown both theoretically and experimentally that there are in fact numerous ways a single head 2-layer transformer can learn Dyck languages: the second attention block needs only satisfy a mild condition for this. We corroborate this finding with our in-depth study of separation accuracy of attention heads.

\section{Problem Statement}
\label{sec:problem}



In this paper we target a simple counting problem where the model has to
decide if the input contains more '1' tokens than '0' tokens. The input may
contain an arbitrary number of '2' tokens interspersed between the '0's and
'1's (which may also occur in any order). For ease of analysis, the input has
an '=' token on the penultimate position, which must be followed by '4' if
there are more 1s than 0s, and '5' otherwise.

\begin{definition}[Count01 language]   
  
  \label{def:count01} Each string $s$ of the \emph{Count01} language fits the
  regular expression {\tt [BOS]\{0,1,2\}$^*$=\{4,5\}[EOS]} so that the final
  letter is 4 if the substring before the '=' contains more 1s than 0s and 5
  otherwise.  

%
\end{definition}

The choice between '4' and '5' is the {\it semantic} problem of counting, getting
the final '[EOS]' token is the {\it syntactic}
problem. Both tasks are well handled by the (skip)trigram mechanism posited by
\cite{elhage2021mathematical}. Remarkably, it is only the syntax that brings
in cooperation among the attention heads, the semantics shows ensemble
behavior. 

We generate a dataset with 7000 training, 1500 validation and 1500
test sentences as follows: For each sentence, we randomly draw the
number of '0', '1' and '2' tokens separately, then generate the
appropriate number of these tokens in random order. Note that
shuffling does not matter for our present architecture as it does not
include positional embedding. The numbers of '0' and '1' tokens are
drawn from the intervals $[0, 100]$, $[101, 150]$ and $[151, 200]$ for
the training, validation, and test sets respectively. The numbers of
'2' tokens are drawn from the intervals $[0, 100]$, $[0, 150]$ and
$[0, 200]$ for the training, validation, and test sets respectively.

We use this dataset to train a single layer, attention-only generative
transformer language model. For each sample, the model is optimised to
minimise the negative log likelihood of the text after the '=' token. In other
words, the model has to learn to generate one of the '4' or '5' tokens,
followed by the '[EOS]' token indicating the end of the sentence. The
superficial syntax, that the '=' token must be
followed by '4' or '5' is learned early in the training, but this permits only
50\% correct (in other words, random) recognition before the counting aspect
is learned. 

\section{Understanding Attention}
\label{sec:minimal_solution}

We briefly describe the computation performed by a transformer
attention layer, which is the primary mechanism for interaction among
tokens. Its input is a sequence of token embeddings $\{e_i \mid 0 \leq
i < l\}$
where $l$ is the length of the sequence and each token
embedding is a learned d-dimensional vector, i.e., $e_i \in
\mathbb{R}^d$.

An optional normalization layer ensures that each embedding vector has learned
mean $\mu$ and variance $\sigma$:
\begin{align*}
  e_i & := \frac{e_i - \text{mean}(e_i)} {\text{std}(e_i)} \sigma + \mu
\end{align*}

The embedding sequence is processed by a number $a$ of attention
heads. Each head produces contextual token embedding vectors.
%

Each head has an internal dimension $d_0$, which usually equals $d/a$.  Each
head $h$ ($0 \leq h < a$) has learned $d \times d_0$ dimensional \emph{Key}
($K_h$), \emph{Query} ($Q_h$), and \emph{Value} ($V_h$) parameter matrices.
Applying these matrices to the token embeddings $e_i$ produces $l \times d_0$
dimensional \emph{key}, \emph{query}, \emph{value} feature matrices $k_h$, $q_h$,
$v_h$.


The key and query feature matrices together create an $l \times l$
dimensional \emph{attention weight} matrix $A_h$ where $A_{h,ij}$
specifies how strongly the $j$th token influences the output of the
$i$th token. This is computed by taking the scalar product of the
query and key feature vectors, obtaining the \emph{attention logits}
and then normalising each row to sum to 1:
\begin{align*}
  A^0_h & = q_h \cdot k_h^T \\
  A_{h,i} & = \text{softmax}(A^0_{h,i})
\end{align*}

\vspace*{-3mm}The attention logit of a particular token pair does not depend on
the input sequence, only on the learned key and query matrices. This
does not hold for the attention value, due to the normalisation
constant in the softmax function. However, if we consider the
attention at some token position $k$ to two tokens $i$ and $j$, then
their ratio $\frac{A_{h,ki}}{A_{h,kj}}$ is input independent since the
normalisation constant is cancelled out. We will exploit this fact to
characterise heads by attention weight ratios of some token types.

The head output is the sum of the head values, weighted by the
corresponding attention vector:
\begin{align*}
  O_h & = A_h \cdot v_h
\end{align*}
Since the attention vector for each token has nonnegative entries that sum to 1, the output is a
convex combination of the value vectors. The concatenation of the outputs of all heads are
projected back to $d$ dimensional vectors (via another learned linear
transformation $P$), dropout is applied and then the output is added to the input embedding.

The attention layer may be followed by a tokenwise small MLP, again preceded by layer normalization and followed by dropout,
constituting a single transformer block. Any number of transformer
blocks could be stacked on top of each other, but in this paper we
focus on single layer, attention-only models.

Moreover, we fuse the linear transformation $P$ into the linear trasformation in the output layer. We can do this as we noticed that, just like in \citep[\S5 and Appendix A.1]{Nanda:2023}, the skip connection has negligible effect.
The output layer of the model is a tokenwise affine transformation
with weights $w_o$ and bias $b_o$ that produces logits for each
possible token type.

\subsection{A Minimal Solution}\label{ss:minimal}

The Count01 language is simple enough so that a very small single layer,
attention-only transformer can solve it with or without layer norm. Given a
partial input sequence, e.g., '\verb|000010=|', the model has to perform two
tasks:

\begin{compactenum}
  \item Whenever it sees the '=' token, it has to count '0's and '1's
    and output either '5' if there are more '1's or '4'
    otherwise.
  \item Whenever it sees a '4' or '5', it has to output an '[EOS]'
    token, indicating the end of the sentence.
\end{compactenum}

It turns out that a single one-dimensional head ($a=d=1$) is sufficient
to correctly count '0's and '1's. The head has to learn an attention
pattern in which the '=' token attends to the '0' and '1'
tokens with different but large weights. Based on their relative weights, a linear 
separator is then sufficient to decide what to write after the '=' sign.
The second task is also simple. The head learns to attend '4' and '5' tokens to themselves 
with an even larger weight, thus the linear separator can predict the '[EOS]' token if the 
head output is large enough.
In Appendix~\ref{app:minsolution} we provide a possible parametrisation of this architecture.

Note, however, that there is a gap between what an architecture can
represent and what it can learn. As we shall see, it is rather
challenging to successfully train the minimal model. To better
understand training dynamics, we continue our investigation with a
somewhat larger, while still small model that trains to high accuracy
consistently.

\section{Understanding Heads}
\label{sec:trained_interpretation}

In this section we delve into the capabilities and dynamics of
individual heads. We focus on predicting the token after '='. See Subsection \ref{ss:eos} for the syntactic task of outputting an '[EOS]' token after '4' or '5'.

First of all, let us introduce some notation for predicting the token after '='.
Given head outputs $\{O_i \mid 0 \leq i < a\}$, the output layer has
weight vectors $\{w_{i,t} \mid 0 \leq i < a\}$ and bias vector $b_t$
for each token $t$. The logit $z_t$ for token $t$ is computed as
$\sum_{i=0}^a O_i^T w_{i,t} + b_t$. Note that this means head $i$ makes a contribution of $z_{t, i}:=O_i^T w_{i,t}$ to this logit.

The model correctly solves the
counting task if $z_4 > z_5$ whenever there are more 1's in the input
than 0's.\footnote{We are here disregarding the fact that the model
could in principle select a token other than '4' or '5', because this
never happens after the first few steps of training.} In other words,
the model's decision depends on the sign of $z_4-z_5 = \sum_{i=0}^a
O_i^T (w_{i,5} -w_{i,4})$.

\subsection{Three Metrics: l-acc, ROC AUC, and s-acc}

Now we shall introduce three metrics to measure the performance of a
group of heads $H \subseteq\{0,\dotsc,a-1\}$.

\begin{figure*}[ht]
  \begin{center}
    \includegraphics[width=0.98\linewidth]{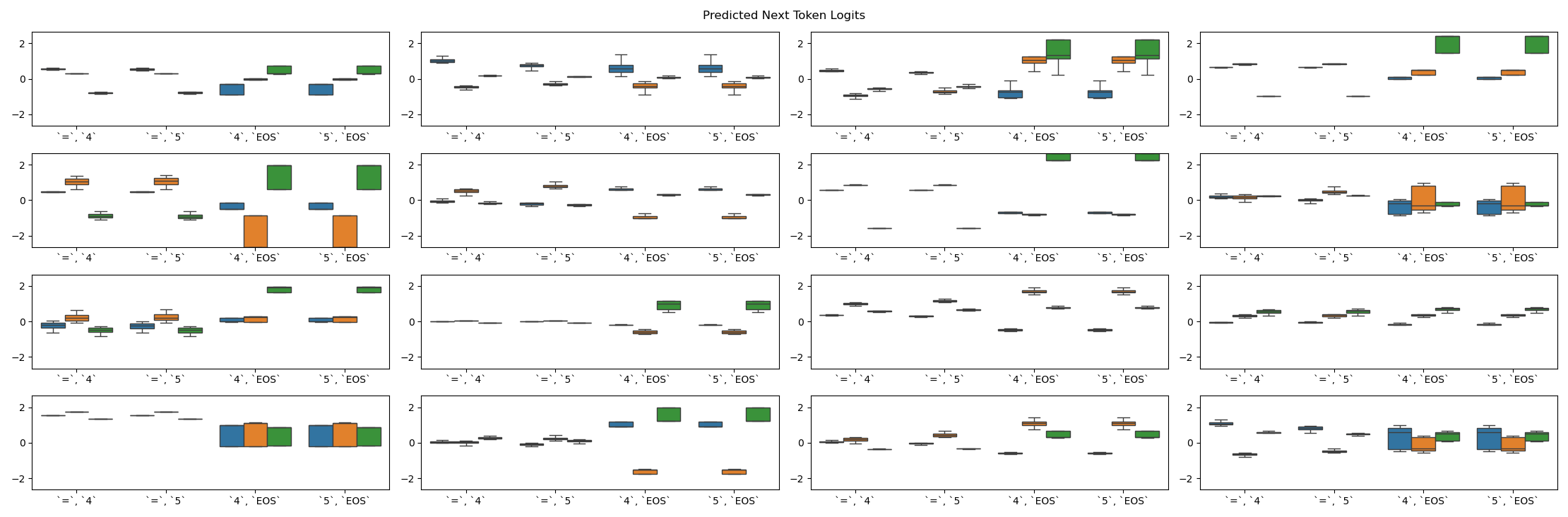}
  \caption{Next token logit distributions for '4', '5', and '[EOS]', separated by current and true next tokens.}
  \label{fig:next token logits}
  \end{center}
\end{figure*}

\emph{Learned Accuracy
(l-acc)} of $H$ is the model accuracy on the test set when all heads
outside $H$ are ablated, i.e., their output is zeroed out. That is, this is the accuracy of the predictions we get with logit for token $t$ given as $z_{t, H}:=\sum_{i\in H}z_{t,i} + b_t$. See Figure \ref{fig:next token logits} for logits given by single heads. For now, let us concentrate on '4' and '5' logits at '='.

One can see on Figure \ref{fig:next token logits} that many heads are highly biased toward one of '4' or '5'. This is what makes the cooperation of heads a central topic in our paper. On the other hand, one can also see that most heads make a meaningful contribution, to the final goal of getting $z_4 > z_5$ if and only if there are more 1's in the input than 0's. The second metric measures this contribution: we can treat either $z_{4, H}$ or $z_{5, H}$ by itself a binary classification logit in the task of determining whether the next token is '4' resp.~'5' or not. Then for these two binary classification problems, we can calculate the ROC AUC values. In our setting, the \emph{ROC AUC} metric is the maximum of these two values.

The third metric exploits the fact that the final logits are computed
from the attention heads by an affine
transformation, i.e, the usefulness of heads -- or groups of heads --
can be measured by their ability to linearly separate the inputs into
correct classes.  \emph{Separation Accuracy (s-acc)} of $H$ is
computed by fitting a linear separator on the concatenation of the head outputs $O_i$ for $i\in H$
based on the training set and computing accuracy on the test
set. Details about how we identify the linear separator can be found
in Appendix~\ref{app:separator}.

Figure~\ref{fig:unary_size_performance} shows that increasing the
number of heads and the size of the embedding dimension makes learning
more robust. We pick one stable setup for most of our experiments: 32
embedding dimensions with 16 heads, which almost always reaches near
perfect test accuracy. Conveniently, in this setup the head dimension
is $2$, making visualization easier.

\begin{figure*}[ht]
  \begin{center}
    \includegraphics[width=0.98\linewidth]{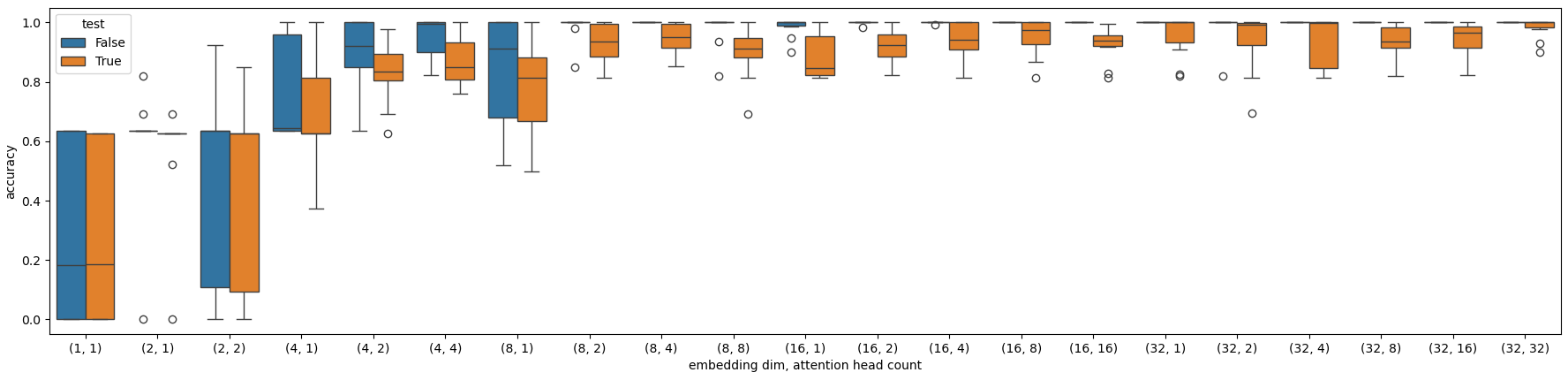}
  \caption{Average model accuracy on the Count01 task, along with standard deviation based on 10 random seeds.}
  \label{fig:unary_size_performance}
  \end{center}
\end{figure*}

\subsection{Understanding Individual Heads}

\begin{figure*}[ht]
  \begin{center}
    \includegraphics[width=0.98\linewidth]{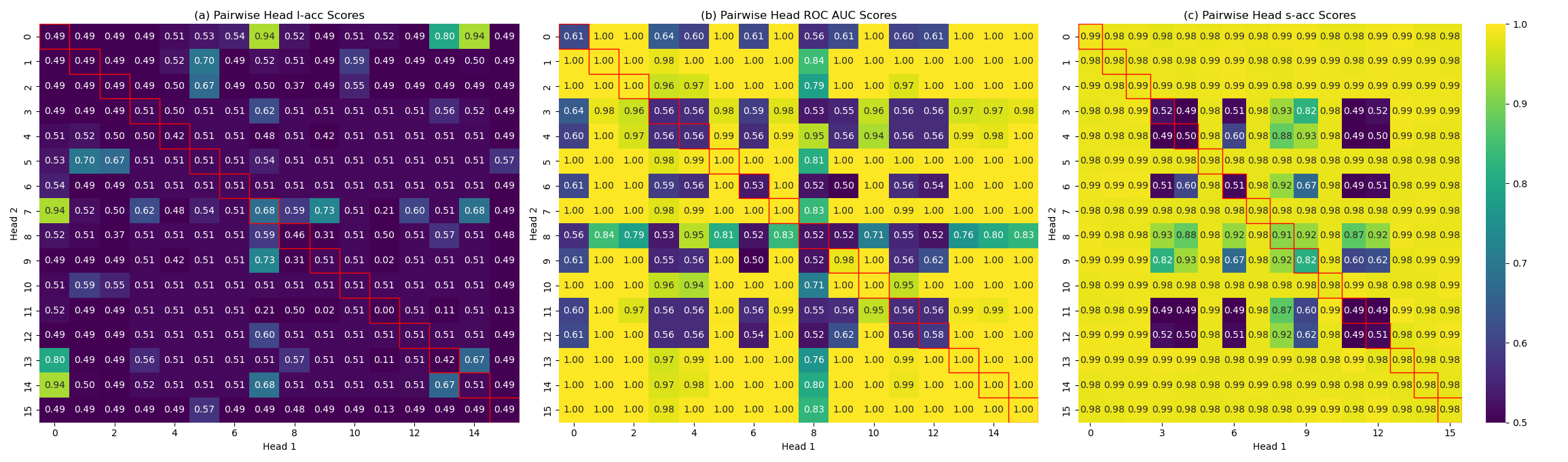}
  \caption{Single and dual head (a) l-acc, (b) ROC AUC, and (c) s-acc on the test set.}
  \label{fig:pairwise scores}
  \end{center}
\end{figure*}

We turn to analysing heads in isolation.  At the '=' token, the
model is expected to linearly separate inputs with more '1's
(requiring a '4' token) from the rest of the sequences (requiring a
'5' token). Figure \ref{fig:pairwise scores} shows l-acc, ROC AUC, and s-acc values of singleton
heads and head pairs. More plots are available in
Appendix~\ref{app:head_pairs}.

Based on the s-acc scores, we group the heads into three categories.
Around half of the heads are \emph{successful}, having above $98\%$
accuracy. Around $40\%$ of the heads are \emph{failed}, i.e., their
performance is close to that of coin flip. We call the few remaining
heads \emph{mixed} with mediocre performance.

A very different
picture emerges when considering the l-acc scores. All but one single head have
50\% l-acc score, i.e., they are not better than random guessing and only two head pairs rise above 80\%. What
this shows is that training produces several near perfect heads,
but the model does not learn to directly exploit any of them.

One can see that the ROC AUC scores strike a middle ground. A good ROC AUC score signifies meaningful contribution disregarding biases. However, the model uses these biases to predict the '[EOS]' tokens after '4' or '5': see Subsection \ref{ss:eos}.

\begin{figure}[htb]
  \begin{center}
    \includegraphics[width=0.98\linewidth]{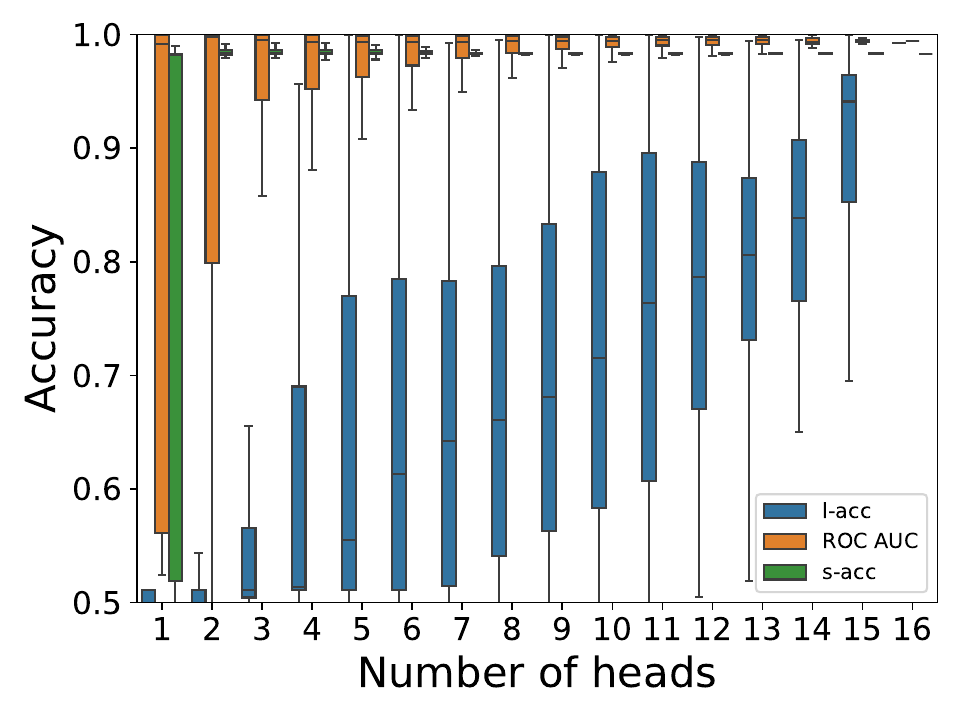}
    \caption{Distribution of s-acc, ROC AUC, and l-acc values of all subsets of heads of a given size.}
    \label{fig:head_groups_barplot}
  \end{center}
\end{figure}

An even stronger demonstration of this pattern emerges if we compute
s-acc, ROC AUC and l-acc scores for each subset of
heads. Figure~\ref{fig:head_groups_barplot}
shows that groups of 3-4
heads already yield close to perfect separation accuracy and ROC AUC values follow close, while most
of the heads are required to achieve strong learned accuracy.

\subsection{The Role of Attention}
\label{ss:role of attention}

We now show that the s-acc performance of a head is determined by its
attention pattern, more precisely how it divides attention among the '0', '1'
and '2' tokens. To understand why attention to other tokens is not important,
note that in the Count01 language, whenever the '=' token is processed, there
is exactly one of the '[BOS]' and '=' tokens, and zero of the '4', '5', and
'[EOS]' tokens. Hence, their contribution to the head output is constant
across inputs, i.e., they do not influence the separability of the samples
regardless of their attention weights.

\begin{figure}[htb]
  \begin{center}
    \includegraphics[width=0.98\linewidth]{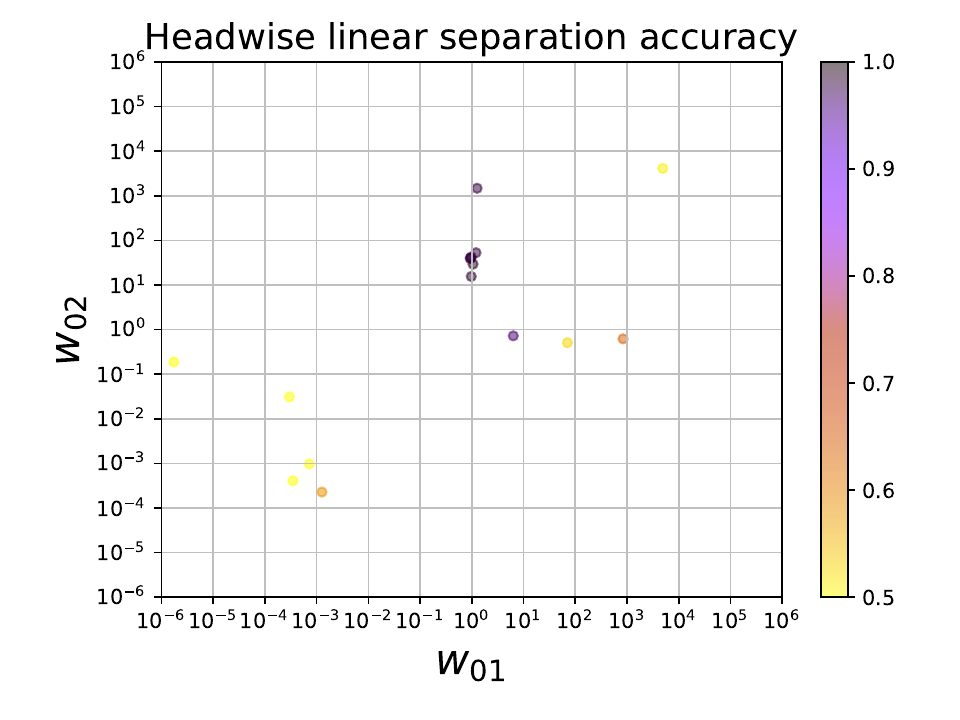}
    \caption{Each dot corresponds to one of the 16 trained heads. Color
      indicates linear separation accuracy. The x axis is $w_{01}$, the
      weight ratio of '0' and '1' tokens. The y axis is $w_{02}$ the
      weight ratio of '0' and '2' tokens.}
    \label{fig:head_performance}
  \end{center}
\end{figure}

As noted in Section~\ref{sec:minimal_solution}, a head can be
characterised by the ratio of two tokens at some position since this
value is input independent. We focus on the attention at the '=' token position
and compute the attention weight ratios among the $0$, $1$ and $2$
tokens.
We use $w_{01}$ to denote the weight ratio between the '0' and '1'
tokens and $w_{02}$ to denote the weight ratio between the '0' and '2'
tokens. Figure~\ref{fig:head_performance}
shows the s-acc of each
head as a function of $w_{01}$ and $w_{02}$. More plots are available in Appendix~\ref{app:head_performance}.
There are two clearly
visible patterns on the figure:

\begin{enumerate}
\item Successful heads have $w_{01} \approx 1$, i.e., the head attends
  roughly equally to the '0' and '1' tokens.
\item Successful heads have $w_{02} > 10$, i.e, token '2' gets much less attention than the '0' and '1' tokens.
\end{enumerate}

\begin{figure}[htb]
  \begin{center}
    \includegraphics[width=0.98\linewidth]{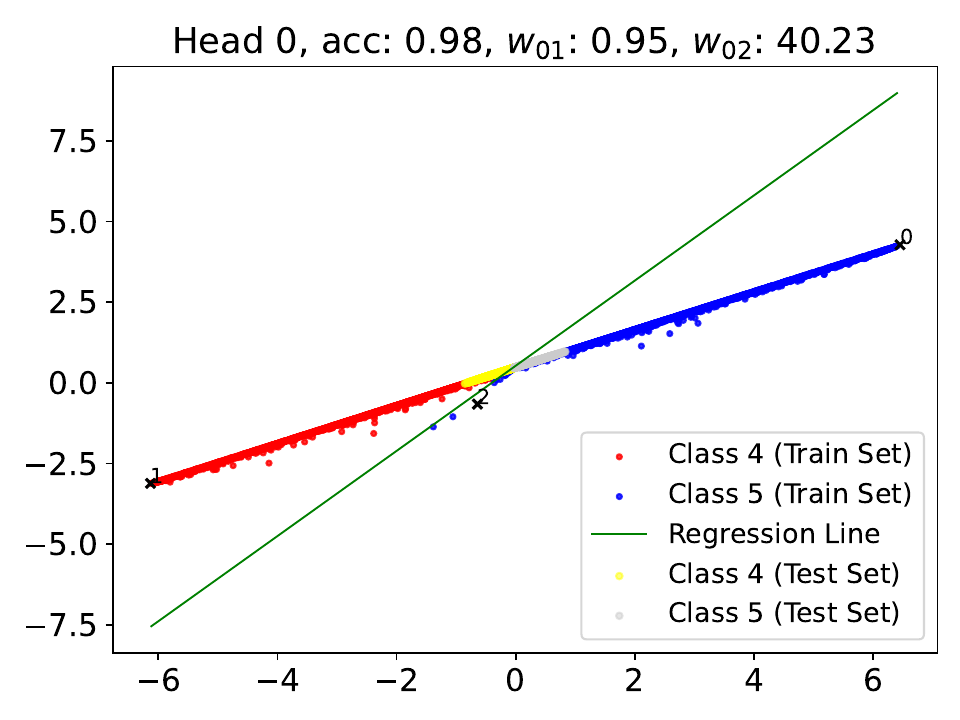}
    
    \includegraphics[width=0.98\linewidth]{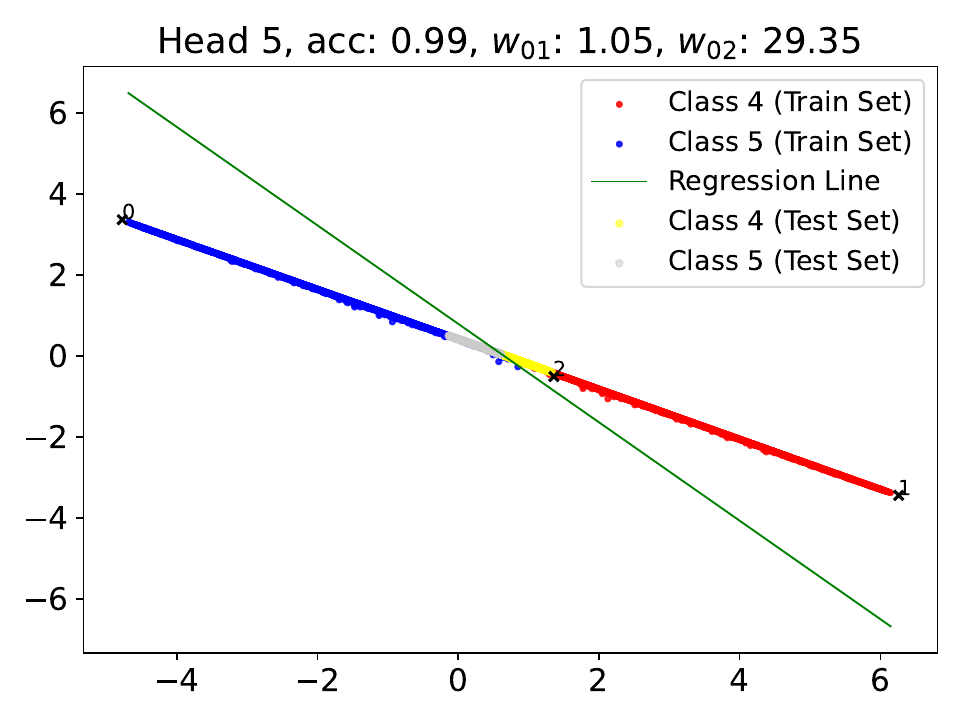}
    \caption{Two successful heads. Colored dots indicate head outputs
      of train and test samples. The green line shows the identified
      linear separator. We also show the value feature vectors of the
      '0', '1', '2' tokens.}
    \label{fig:head_outputs_selection_successful}
  \end{center}
\end{figure}

In order to better understand how attention distribution influences
the heads, we plot the head outputs for the test and train samples,
using different colors to distinguish classes, along with the value
feature vectors associated with the '0', '1', '2' tokens. This is
shown in Figure~\ref{fig:head_outputs_selection_successful} for two
successful heads and in Figure~\ref{fig:head_outputs_selection_failed}
for two failed heads. More plots are available in
Appendix~\ref{app:task1}.  We observe that successful heads indeed
disregard the '2' token ($w_{02}$ is large) and linearly interpolate
between the values of the '0' and '1' tokens ($w_{01} \approx 1$). In
the failed heads, we see two failure modes. First, when the head
attends much more to '0' than to '1' ($w_{01}$ is small or large),
then the prediction collapses into a single point.  Second, when the
head attends to the '2' token ($w_{02}$ is small), it counts those
tokens as well, introducing noise to the prediction.

\begin{figure}[htb]
  \begin{center}
    \includegraphics[width=0.98\linewidth]{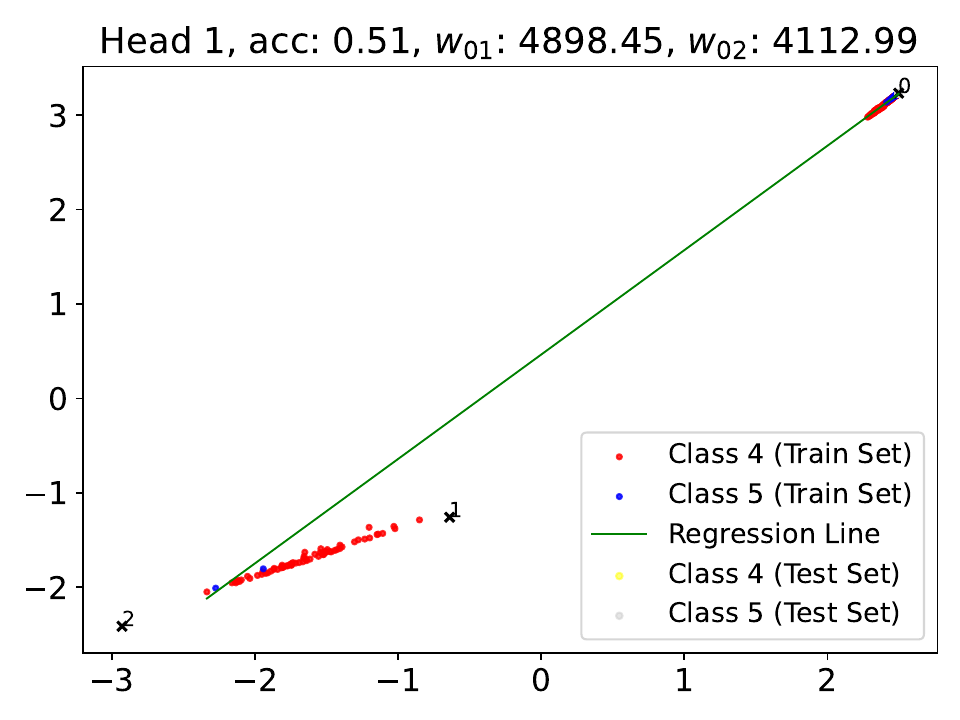}
    
    \includegraphics[width=0.98\linewidth]{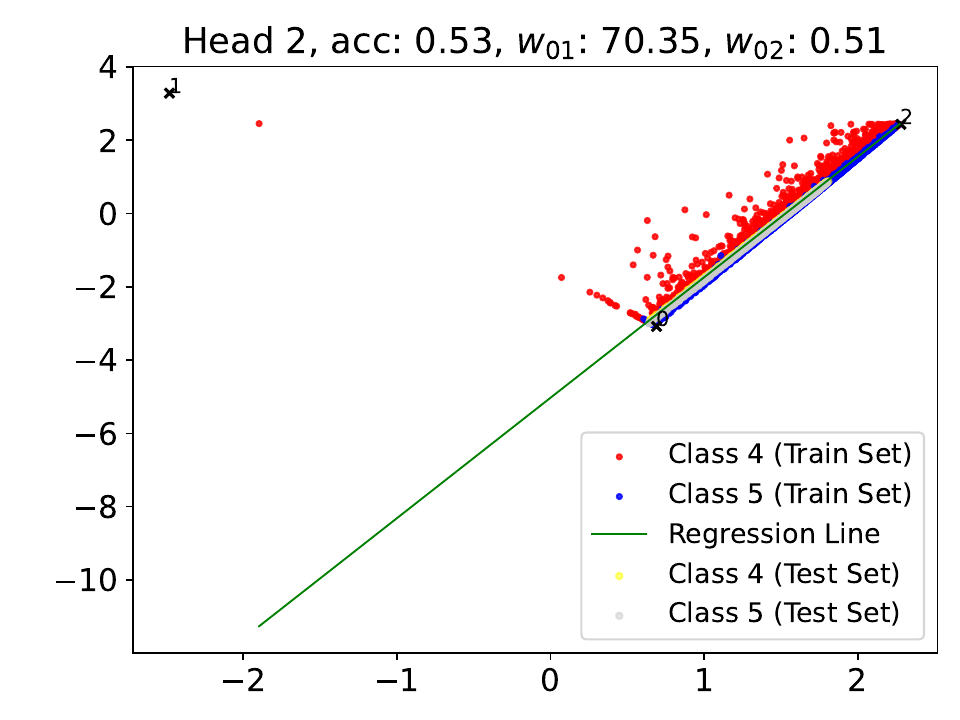}
    \caption{Two failed heads. \textbf{Top}: All predictions collapse
      into a single point. Note, the outlier points between the '1'
      and '2' value vectors: these are 73 training samples with no '0'
      tokens in them, two of which don't even have '1's.
      \textbf{Bottom}: '2' tokens introduce a lot of noise.}
    \label{fig:head_outputs_selection_failed}
  \end{center}
\end{figure}
\subsection{Interventions}

Our investigations suggest that head peformance in the Count01 task is
essentially determined by the attention pattern and the value feature vectors
play minor role. We further highlight this by taking a trained model and make
interventions on the attention matrix, without touching anything
else. We focus on the attention matrix at the '=' token and zero out
all attention weights except for the '0', '1', '2' tokens.

\begin{figure}[htb]
  \begin{center}
    \includegraphics[width=0.98\linewidth]{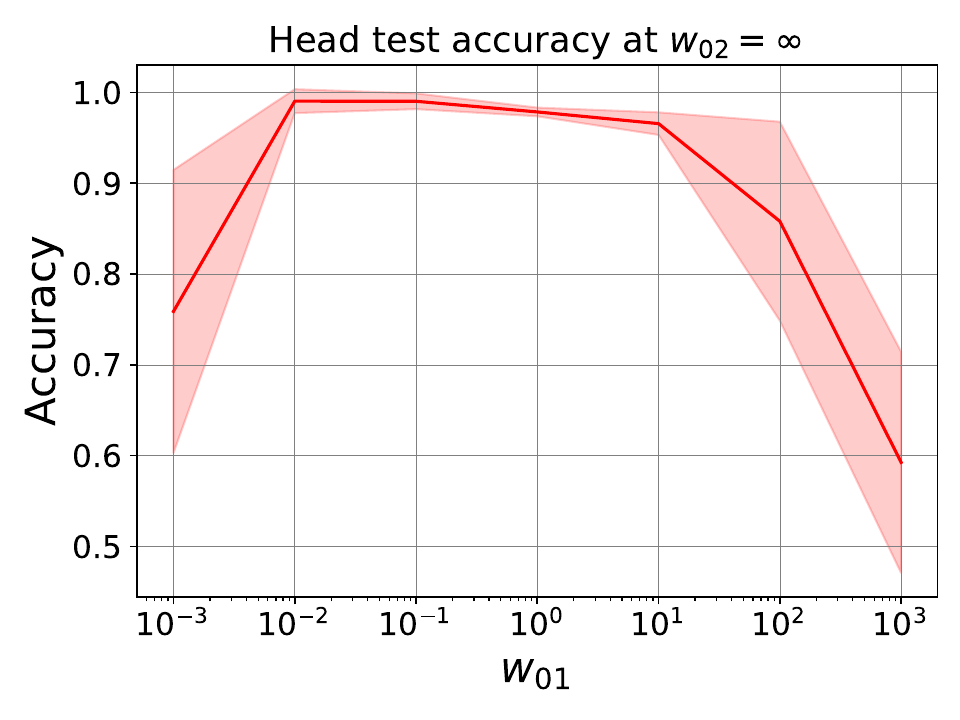}
    \caption{S-acc distribution of heads when the attention matrix is
      manually adjusted. Shaded areas indicate standard deviation.
      Head attention is zero for every token except for '0' and '1' and
      we vary the attention ratio $w_{01}$ between these two tokens.}
    \label{fig:interventions1}
  \end{center}
\end{figure}

Figure~\ref{fig:interventions1} shows head performance distribution
when attention to the '2' token is set to zero and we vary the
attention ratio among the '0' and '1' tokens. All heads perform near
perfectly when this ratio is neither too small not too large (between
0.1 and 10). More extreme ratios yield sharp drop in the average
performance, as well as greater variance. Hence, we conclude that as
long as the '2' token is not attended to and $w_{01}$ is in the
healthy region, other head features, i.e, the value feature vectors barely
influence separability.

\begin{figure}[htb]
  \begin{center}
    \includegraphics[width=0.98\linewidth]{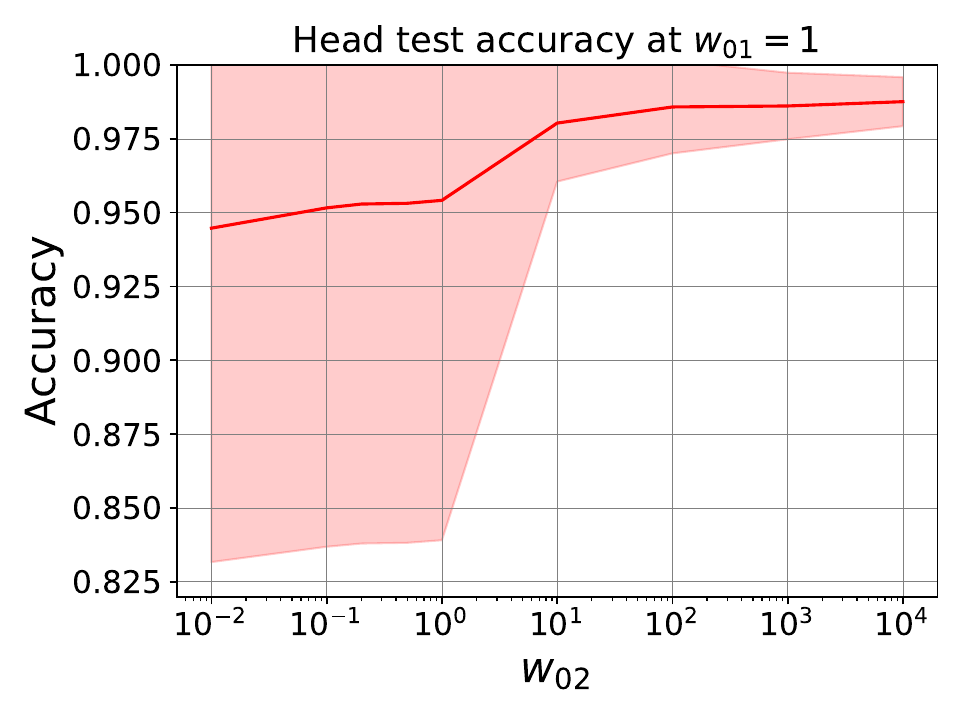}
    \caption{S-acc distribution of heads when the attention matrix is
      manually adjusted. Shaded areas indicate standard deviation.
      Head attention for '0' and '1' are equal and we vary the attention
      ratio $w_{02}$ between the '0' and '2' tokens.}
    \label{fig:interventions2}
  \end{center}
\end{figure}

Figure~\ref{fig:interventions2}
shows that changing the relative
attention assigned to the '2' token has a huge impact on
performance. The greater the $w_{02}$ ratio, i.e., the less attention
'2' gets, the better the performance and the less variance among the
trained heads.

\subsection{Head Evolution During Training}

As we have seen in Figure~\ref{fig:pairwise scores}c, training results
in roughly half of the heads being successful and capable of strong
linear separation. We now look at the evolution of s-acc scores during
training. Figure~\ref{fig:sacc_evolution} in
Appendix~\ref{app:sacc_evolution} reveals that heads initially
manifest all sorts of s-acc values, including extremely strong
ones. In the initial phases of training s-acc values change a lot and
we see extreme transitions, i.e., very bad heads becoming very good
and vice versa. In around 50 epochs the heads start to stabilise and
we obtain a heavily bipolar distribution: the trained heads are either
very good or very bad. In the figure, line width indicates the weight
in the output layer associated with the head. As to why the
weights barely change at all during training, that is the model does not learn
to focus on the good heads, see Subsection \ref{ss:eos}.

\section{Interaction among Heads}

We now turn our attention to interaction among
heads. Figure~\ref{fig:pairwise scores}c shows how pairs of heads
improve or impair the linear separation capability of individual
heads. Improvement is rare and it mostly happens when two failed heads
are considered together, but they still perform poorly. Nevertheless,
in each experiment we do see rare examples of a successful head made
even better by another head, even reaching perfect separation
accuracy, see e.g. heads (5,13) in Figure~\ref{fig:pairwise scores}c.

We have seen previously in Figure~\ref{fig:pairwise scores}a that the
model does not learn to fully exploit successful heads. How about
pairs of heads? The situation is not fundamentally different, as we
see that learned accuracy scores for pairs of heads are almost always
around 50\%. Even if there are exceptions,
Figure~\ref{fig:head_groups_barplot} shows that a large number of
heads is required to achieve large learned accuracy and all the 16
heads are required to achieve perfect l-acc score.


\subsection{Head Cooperation}

Is there any cooperation among heads? Is a trained model more than an
ensemble of heads? 
One can easily construct solutions to the Count01 problem in which
heads actively cooperate. For example, there could be one head that
counts '0's, another that counts '1's and the output layer could
simply compare the outputs of these heads. Alternatively, the number
of '2' tokens could be split into intervals and each head could learn
to separate well in one of the intervals. In such cooperative
solutions, we should see pairs of heads performing better than
individual heads. However, Figure \ref{fig:pairwise scores}c shows that such positive interactions are rare and the s-acc values of single heads are already high.

If the model is indeed an ensemble of heads, then we should be able to
predict the model's performance from that of the heads. We now show
that such prediction is quite possible.

\begin{figure}[htb!]
  \begin{center}
    \includegraphics[width=0.98\linewidth]{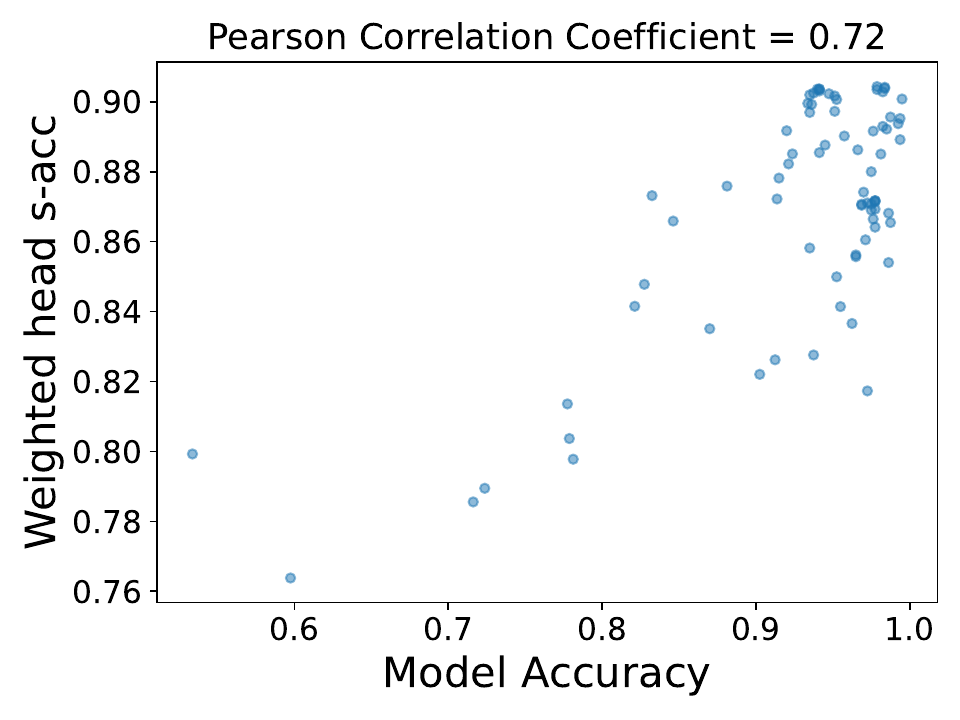}
  \caption{Scatter plot showing the correlation among weighted head
    s-acc values and model accuracy. Each dot corresponds to a
    checkpoint between the 10th and 100th epoch from four experiments
    with different initialisations.}
  \label{fig:correlation}
  \end{center}
\end{figure}

For what follows, recall the notation introduced in the beginning of Section \ref{sec:trained_interpretation} We compute the lengths of the $w_{i,5}
-w_{i,4}$ vectors and normalise them across heads to sum to 1. We call
the obtained values the model learned \emph{head weights ($hw$)}. We
show that if we compute the weighted sum of the s-acc values, with
$hw$ used as weights, then we get a value that very strongly
correlates with model accuracy. This can be seen in the scatter plot
of Figure~\ref{fig:correlation}, which shows that the Pearson
correlation coefficient of the two values is $0.72$. This strong
correlation suggests that it is almost the same if 1) we first compute
separation accuracy for each head and then average them weighted
according to the final model or 2) we first compute the model output
from the heads by the final layer and then compute separation
accuracy. This suggests that the model is working as a proper ensemble,
solving two separate tasks: 1) make the heads as good as possible and
2) learn a good linear combination of the head outputs.

\subsection{The Syntactic Task: Predicting the '[EOS]' Token}\label{ss:eos}
While our analysis concentrates on the counting task, i.e, deciding
whether there are more '1's than '0's in the input, we briefly mention
the second, more technical task that the model has to learn: to put an
'[EOS]' token after the answer. This is an extremely simple task and
the model requires minimal capacity to learn it: based on 5
experiments (using 80 heads altogether), we find that heads and head
pairs reach 100\% learned accuracy with probability 0.46 and 0.5,
respectively.

As to how the '[EOS]' token logits coalesce: we already noticed in Section \ref{sec:trained_interpretation} how the heads are typically biased toward one out of '4' or '5'. See Figure \ref{fig:next token logits} in Appendix \ref{app:task1} for next token logit distributions of individual heads for all of '4', '5', '[EOS]', over both '=', and '4' or '5'. One can see that in a single head, the ordering of the logits is about fixed, but the balance between the logits tips toward '[EOS]' when we move from '=' to '4' or '5'.

\section{Discussion and Conclusion}

In this work, we use mechanistic interpretability to study how single-layer attention-only transformers solve a simple counting task. We notice that although there exists a very small minimal solution, in order for randomly initialized models to reliably learn to solve the task, we need to make them somewhat bigger.

We focus on an architecture with 16 attention heads, thus giving us opportunity to study the orchestration between heads. We show that they individually learn to solve the main task, but their outputs need to be aggregated in a non-uniform way so as to solve the auxiliary task.

The methods we use are by no means limited in scope to counting tasks: we believe they could serve as tools to mechanistically interpret transformer models trained to other, more complicated tasks.

\newpage

\section*{Acknowledgments}

This work has been supported by Hungarian National Excellence Grant
2018-1.2.1-NKP-00008, the Hungarian Artificial Intelligence National
Laboratory (RRF-2.3.1-21-2022-00004) and the ELTE TKP 2021-NKTA-62
funding scheme.

\section*{Impact Statement}
This paper presents work whose goal is to advance the interpretability
of of Machine Learning models. This line of research has great
potential in making models safer and more suitable for critical
environments. However, our work focuses on small models and simple
tasks and there are no direct consequences which we feel must be
specifically highlighted here.




\bibliography{formulaembedding}
\bibliographystyle{arxiv}

\newpage
\appendix
\onecolumn

\section{A Minimal Solution}
\label{app:minsolution}

Let $n_0$ and $n_1$ be the number of '0' and '1' tokens in the input, respectively. If we have $d=1,d_0=1$, it is possible to construct a Transformer that solves the problem, if $n_0+n_1>0$.
Below, we present a minimal setup that solves the problem. We omit the head indices, since there is only one head. Layer normalisation is not applied, because it would produce the same embedding vector for all tokens if $d=1$. The one-dimensional token embeddings will be

\[
W_E = 
\begin{pmatrix}
    0  \\
    N  \\
    N+1  \\
    0  \\
    1  \\
    N^2  \\
    N^2  \\
    0  \\
\end{pmatrix}
\hspace{0.1em}
\begin{array}{c}
    \text{'[BOS]'} \\
    \text{'0'} \\
    \text{'1'} \\
    \text{'2'} \\
    \text{'='} \\
    \text{'4'} \\
    \text{'5'} \\
    \text{'[EOS]'} \\
\end{array}
\]

and the key, query and value parameter matrices will be $K=Q=V=1$. We are only interested in query feature matrices of tokens '=', '4' and '5', so the relevant token pair attentions logits will be the following:
\SetTblrInner{rowsep=0pt}
$$\begin{tblr}{c|cccccc}
    & k_{BOS}=0 & k_0=N & k_1=N+1 & k_2=0 & k_==1 & k_4=k_5=N^2 \\
    \hline
    q_==1 & 0 & N & N+1 & 0 & 1 & -  \\
    q_4=q_5=N^2 & 0 & N^3 & N^3+N^2 & 0 & N^2 & N^4  \\
\end{tblr}$$
The attention logits of $q_=\cdot k_4$ and $q_=\cdot k_5$ will not be used at all, because '4' and '5' tokens come only after '=' token. For this reason we denoted its values with '$-$'.

Since we take a softmax function for each query vector, therefore if the input ends with '=', then only $A_{=,0}$ or $A_{=,1}$ will be significant. If the input ends with '4' or '5', then $A_{4,4}$ or $A_{5,5}$ will be significant only. Therefore, if $n_0+n_1>0$, then the output of the attention mechanism in the first task will be:
\begin{align*}
    \textbf{y}_=&=\frac{v_{BOS}+n_0e^Nv_0+n_1e^{N+1}v_1+n_2v_2+e\cdot v_=}{1+n_0e^N+n_1e^{N+1}+n_2+e}
    =\frac{e^N(n_0N+n_1e(N+1))+e}{1+n_0e^N+n_1e^{N+1}+n_2+e}
    &\approx\frac{n_0N+n_1e(N+1)}{n_0+n_1e}.
\end{align*}
Let $a$ be defined such as $\textbf{y}_=\ge a \Leftrightarrow n_0\ge n_1$. We get that $a:=\frac{e(n+1)+n}{1+e}$.
Without loss of generality we can assume that the last token is '4' in the second task. Then the output will be 
\begin{align*}
    \textbf{y}_4&=\frac{v_{BOS}+n_0e^Nv_0+n_1e^{N+1}v_1+n_2v_2+e\cdot v_=+e^{N^4}v_4}{+n_0e^N+n_1e^{N+1}+n_2+e+e^{N^4}v_4}\approx \frac{e^{N^4}N^2}{e^{N^4}}=N^2.
\end{align*}
These are only approximations, and are only valid if $n_2\ll e^N$, where $n_2$ is the number of '2' tokens in the input.
However, if we assume that the Transformer is not infinitely precise then with an appropriate large $N$ value, the self attention mechanism outputs these values. Now, we need to define the output layer, and prove that it will produce the correct logits:
\[
\begin{aligned}
    W_{out} &= \begin{pmatrix}
        0 & 0 & 0 & 0 & 0 & 0 & 1 & -1 & 4
    \end{pmatrix} &
    b_{out} &= \begin{pmatrix}
        0 & 0 & 0 & 0 & 0 & 0 & -a-1 & a+1+\varepsilon & -4\cdot 3(N+1)
    \end{pmatrix}
\end{aligned}
\]
where $\varepsilon$ is a small positive number that compensates the approximation error.
Let $e_{<token>}$ denote the embedding vector corresponding to a given token, which contributes to the residual connection, and let $\mathbf{v}_{<token>}$ represent the corresponding logit vector. The output layer yields the following logits for the first task:
\[
  \big(e_=+\textbf{y}_= \big) W_{out}+b_{out}\approx \left(y_=-a\right)\mathbf{v}_4+\left(a-y_=\right)\mathbf{v}_5+c\cdot \mathbf{v}_{[EOS]},
\]
where $c=4\cdot y_=-12(N+1)<0$, which predicts indeed '4' or '5' depending on whether $n_0/n_1\ge 1$ or not.
In the second task, we have
\[
  \big(e_4+\textbf{y}_4\big) W_{out}+b_{out}\approx \left(4N^2-12(N+1) \right)\mathbf{v}_{[EOS]}+(N^2-a-1) \mathbf{v}_4+(a+1-N^2) \mathbf{v}_5
\]
logits, which predicts '[EOS]' token if $N$ is large enough.

This example assumed the skip connection, but the model weights can be slightly modified to achieve the same results without the skip connection.

\section{Trying to Find Minimal, Perfect Heads via Learning}

We have seen that in the minimal solution, a single head with head
dimension $d_0=1$ is sufficient. In this section, we explore how
feasible it is to find such heads via learning. We run a large number
of experiments with fixed $d_0=1$ in several setups, changing both the
token embedding dimension and the number of heads, and monitor the
number of perfect (s-acc $= 1.0$) or successful (s-acc $\geq 0.98$)
heads that arise by the end of the training. The results can be seen
in Table~\ref{tab:head_performance}.

We find that increasing the token embedding dimension $d$ greatly
increases the chance of finding perfect minimal heads. When $d=32$ we
obtain 17 perfect heads out of 100 single head experiments, while for
$d=1$ none of the heads are perfect out of 650 experiments. This
suggests that the minimal head dimension $d_0$ is reasonably
achievable via learning, while the minimal token embedding dimension
$d$ is beyond the reach of gradient descent.

We previously looked at various ways in which the trained model heads
could cooperate and found little evidence of it. There is, however,
another sort of possible cooperation: they could positively enhance
each other's training trajectories. If this is the case, we should see
heads trained jointly to perform better than those trained in
isolation. However, the last four lines of
Table~\ref{tab:head_performance} suggest the contrary. When $d=1$, we
never find any perfect heads, but looking at heads with over $98\%$
s-acc value, we see that single head experiments succeed twice as often
as 64 head experiments ($0.6\%$ vs $0.3\%$). The results are similar
when $d=2$: single head experiments produce perfect heads 8.5
times as often as multi head experiments ($1.7\%$ vs $0.2\%$). We
conclude that it does not help the heads when they are trained
jointly, they perform much better after separate trainings.

\begin{table}[h]
\centering
\begin{tabular}{@{}cccrrr@{}}
\toprule
\multicolumn{1}{c}{\begin{tabular}[c]{@{}c@{}}Embedding\\dim\end{tabular}} & 
\multicolumn{1}{c}{\begin{tabular}[c]{@{}c@{}}Number of\\exps\end{tabular}} &
\multicolumn{1}{c}{\begin{tabular}[c]{@{}c@{}}Heads per\\exp\end{tabular}} & 
\multicolumn{1}{c}{\begin{tabular}[c]{@{}c@{}}Total\\heads\end{tabular}} & 
\multicolumn{1}{c}{\begin{tabular}[c]{@{}c@{}}Heads with\\s-acc $\geq$ 98\%\end{tabular}} & 
\multicolumn{1}{c}{\begin{tabular}[c]{@{}c@{}}Perfect\\heads\end{tabular}} \\
\midrule
32 & 100 & 1 & 100 & 74\% & 17\% \\
16 & 100 & 1 & 100 & 54\% & 11\% \\
8 & 100 & 1 & 100 & 48\% & 9\% \\
4 & 100 & 1 & 100 & 22\% & 1\% \\
2 & 300 & 1 & 300 & 9.7\% & 1.7\% \\
2 & 100 & 64 & 6,400 & 2.0\% & 0.2\% \\
1 & 650 & 1 & 650 & 0.6\% & 0\% \\
1 & 20 & 64 & 1,920 & 0.3\% & 0\% \\
\bottomrule
\end{tabular}
\caption{Performance of heads based on s-acc scores with fixed minimal
  head dimension $d_0=1$ and different token embedding dimension and
  head number. In these experiments layer normalisation was not
  applied, as it reduces the effective embedding dimension.}
\label{tab:head_performance}
\end{table}

\section{Fitting a Linear Separator to a Set of Heads}
\label{app:separator}

We expect that if $H'\subset H$ than the separation accuracy (s-acc)
of $H$ is larger than that of $H'$, since the model can use more
information. However, this is not completely guaranteed in practice,
as we see in some cases. The linear separator minimizes another
objective than the model, and finding the best separator requires many
steps. Altogether, choosing the right separation algorithm can affect
the results.

We use Support Vector Machines (SVM) with linear
kernel as a linear separator. We use squared hinge loss as the loss
function, and $L_2$ penalty. Formally, the optimization problem is the
following:
$$||w_s||^2+C\left[\frac{1}{n}\sum_{i=1}^n
  \operatorname{max}(0,1-y_i(\textbf{r}w_s+b_s))\right]$$
where $w_s,b_s$ are parameters of the linear separation and
$y_i\in\{-1, 1\}$ represents the true label ('4' or '5', respectively)
of the $i$-th token. To reduce the impact of the penalty, we use
$C=1000$. This ensures that there is a smaller margin between the two
classes.

\section{Learned and Separation Accuracy for Different Initializations}
\label{app:head_pairs}

Here we provide replicas of Figures \ref{fig:pairwise scores}a and
\ref{fig:pairwise scores}c for three different random seeds. We see
very similar patterns of l-acc and s-acc values. Futhermore, in all
experiments, around half of the heads are succesful.

\begin{figure}[htb]
  \begin{center}
    \includegraphics[width=0.32\linewidth]{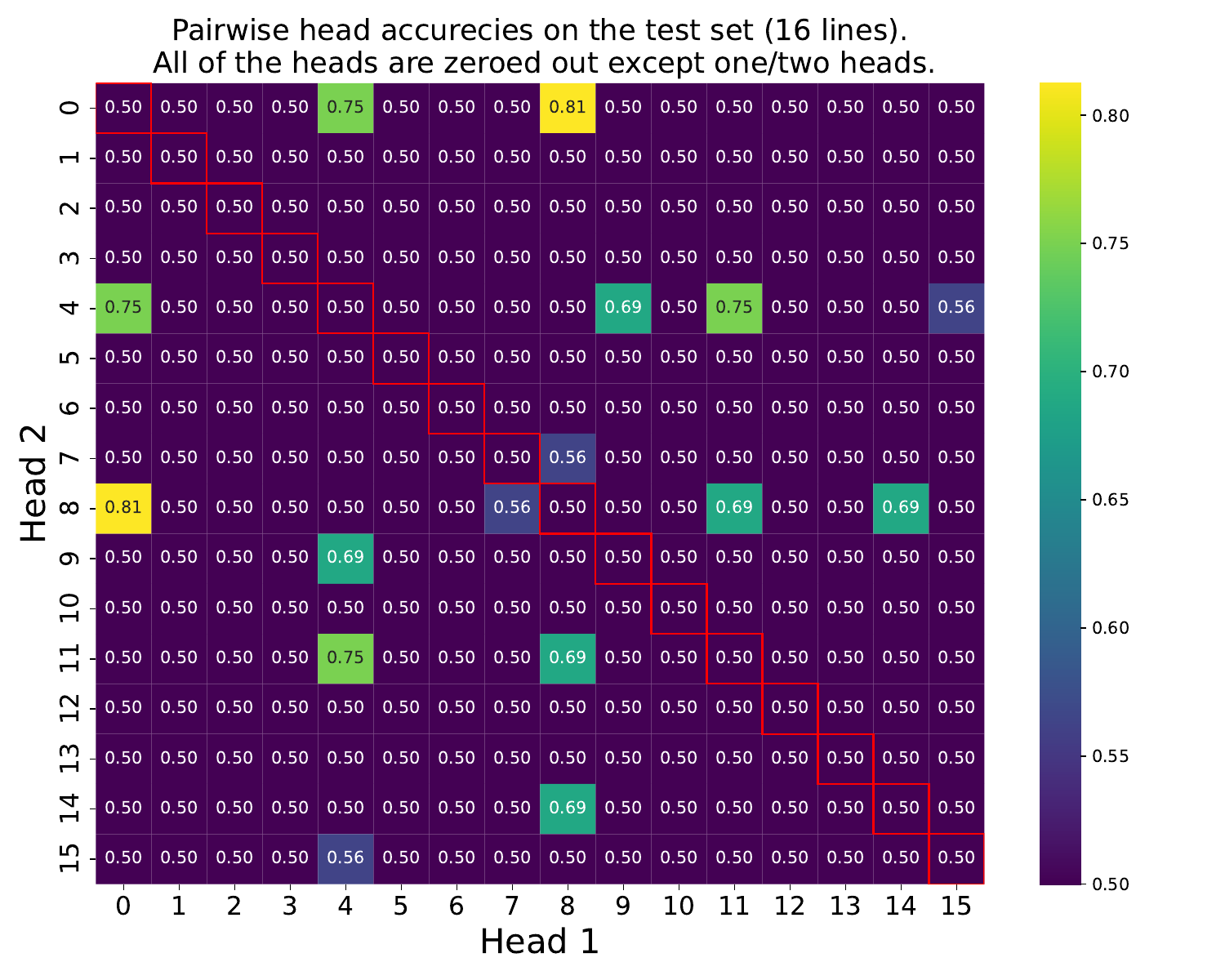}
    \includegraphics[width=0.32\linewidth]{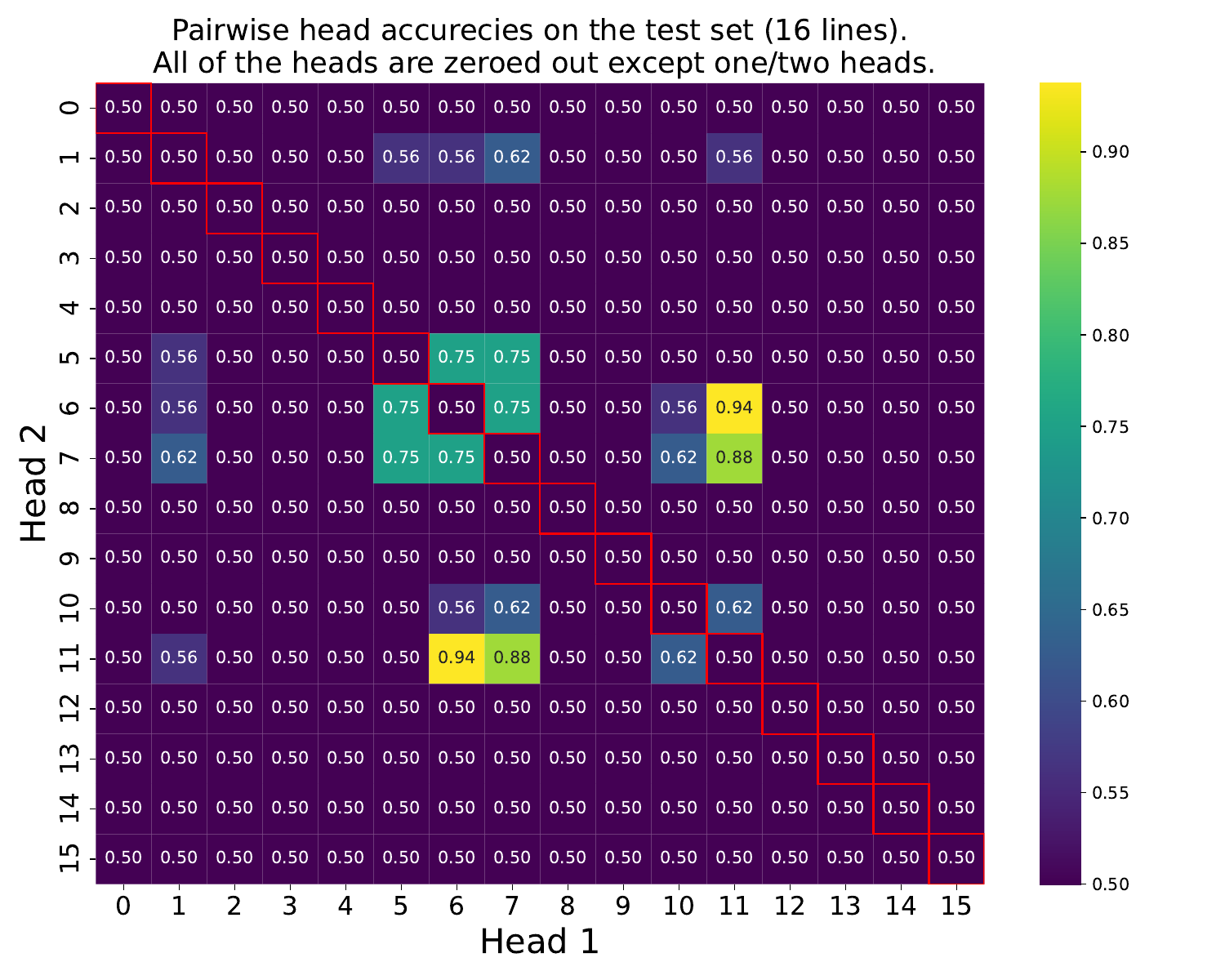}
    \includegraphics[width=0.32\linewidth]{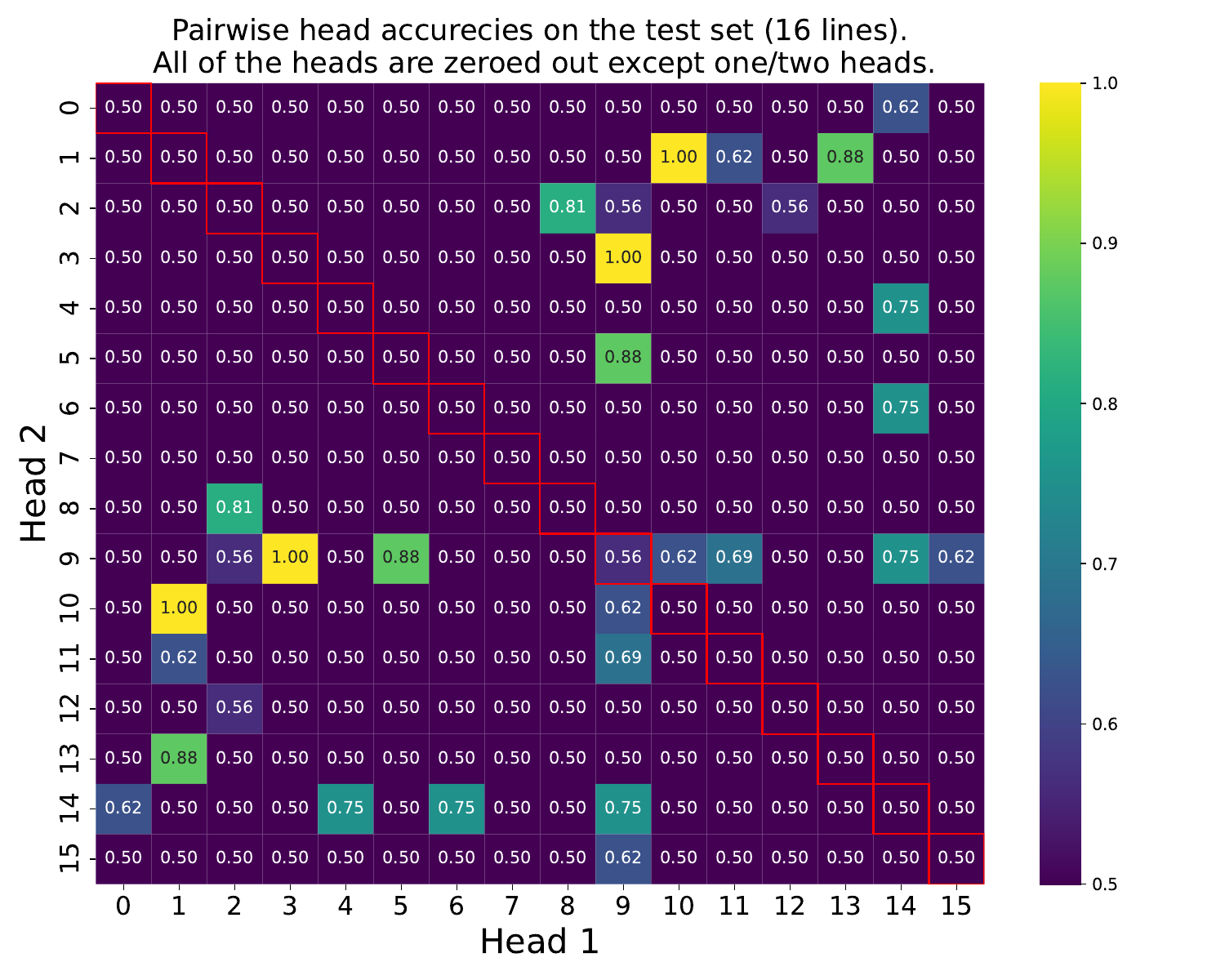}
  \caption{l-acc values of individual heads (in the diagonal) and
    pairs of heads. The plots show results for three different random
    initializations.}
  \label{fig:head_pairs_lacc_x}
  \end{center}
\end{figure}

\begin{figure}[htb]
  \begin{center}
    \includegraphics[width=0.32\linewidth]{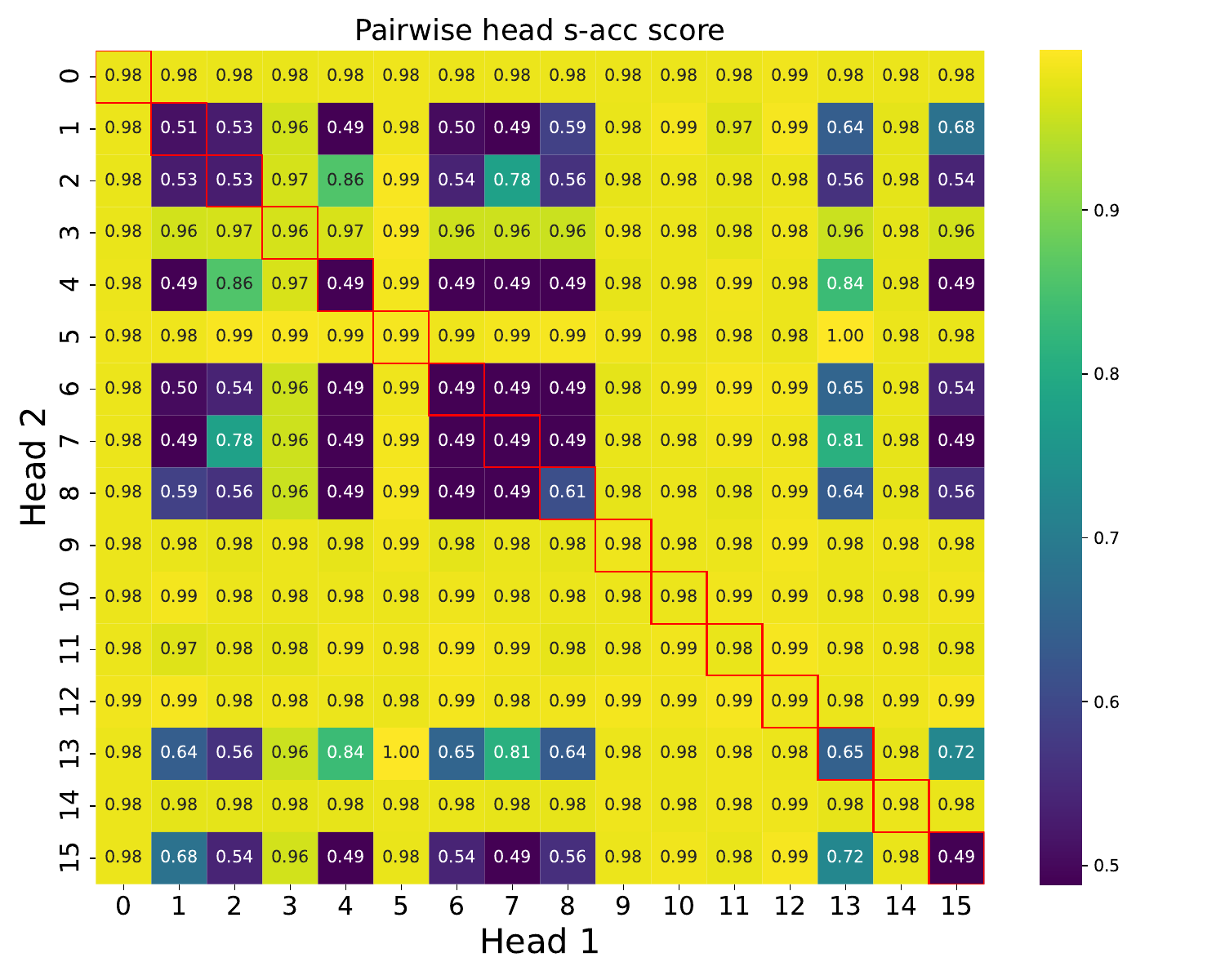}
    \includegraphics[width=0.32\linewidth]{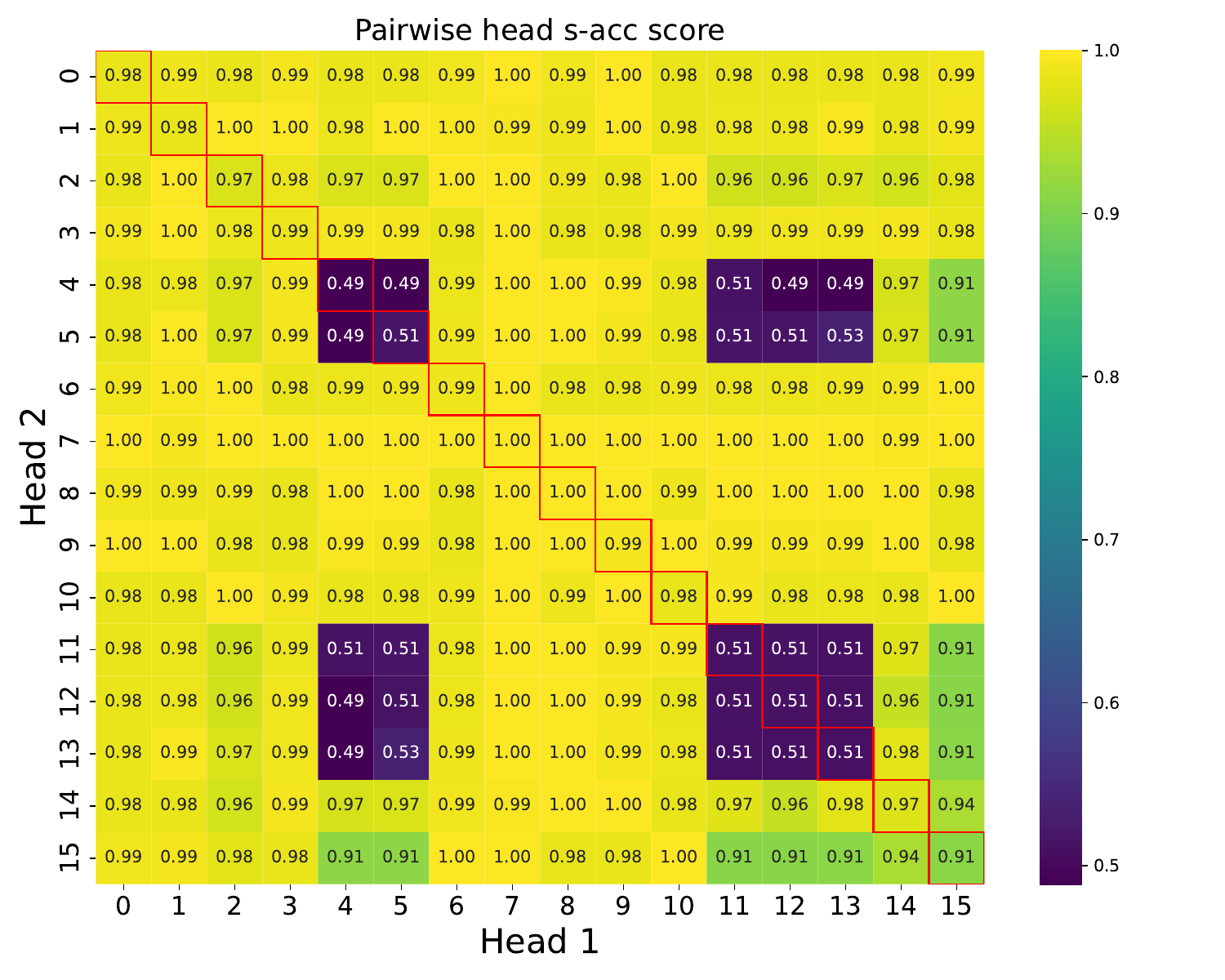}
    \includegraphics[width=0.32\linewidth]{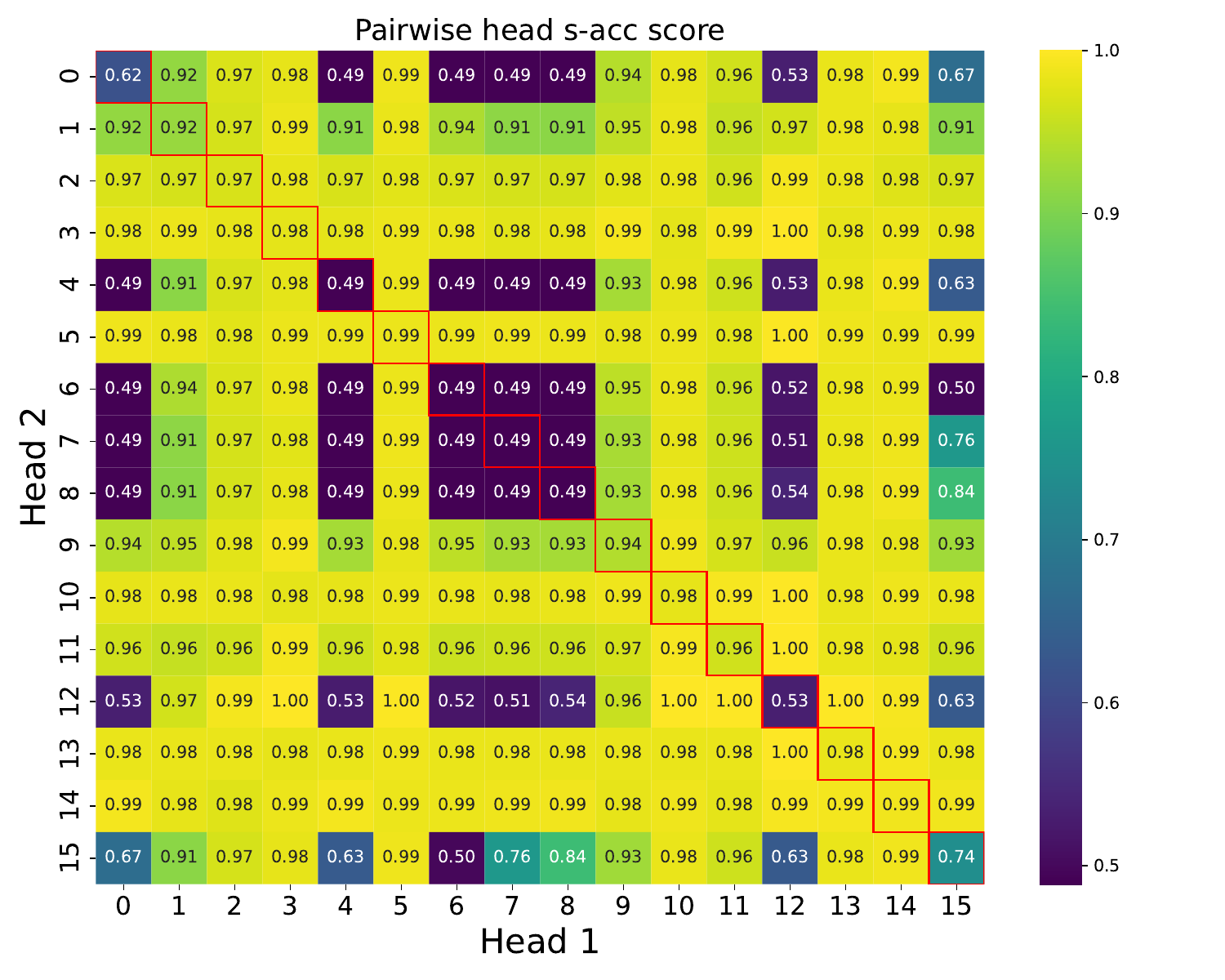}
    \caption{s-acc values of individual heads (in the digonal) and
      l-acc gains of pairs of heads compared to the stronger
      individual head. The plots show results for three different
      random initializations.}
    \label{fig:head_pairs_sacc_x}
  \end{center}
\end{figure}

\section{Separation Accuracy as a Function of Attention}
\label{app:head_performance}

Here we provide replicas of Figure \ref{fig:head_performance} for
three different random seeds. We see very similar patterns: stronger heads have $w_{01} \approx 1$ and $w_{02} > 10$.

\begin{figure}[htb]
  \begin{center}
    \includegraphics[width=0.32\linewidth]{figures/task1_task4.pdf}
    \includegraphics[width=0.32\linewidth]{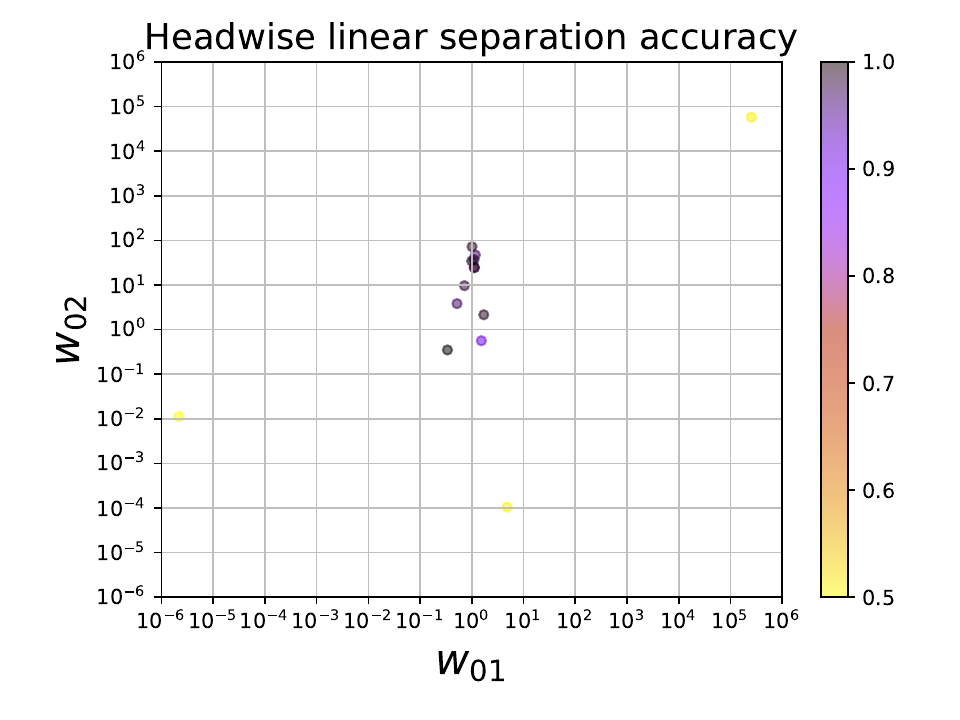}
    \includegraphics[width=0.32\linewidth]{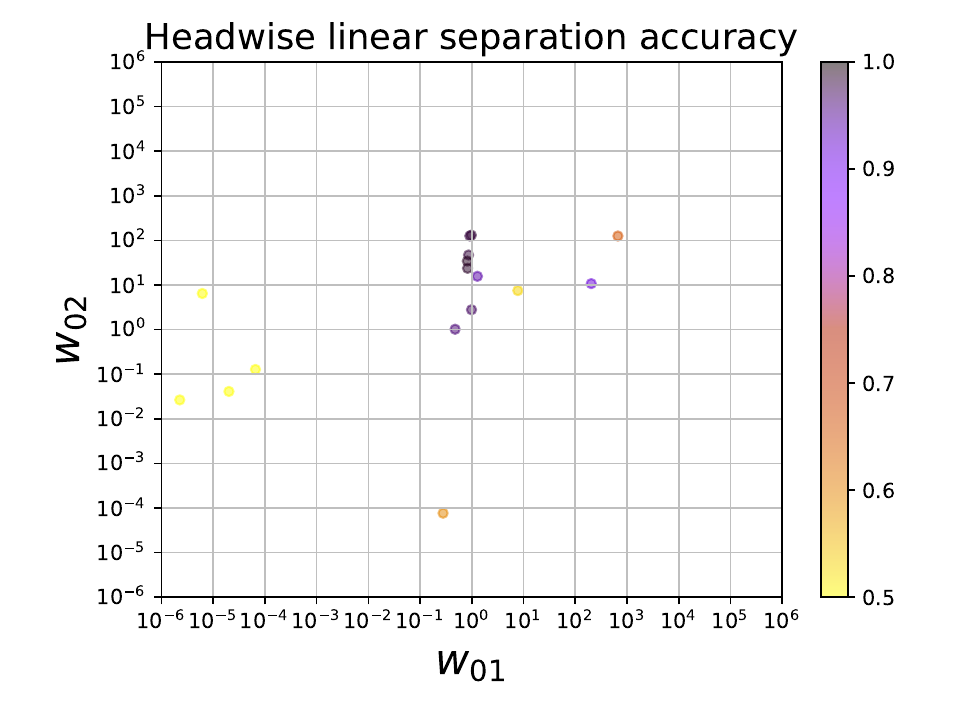}
    \caption{Each dot corresponds to one of the 16 trained heads. Color
      indicates linear separation accuracy. The x axis is $w_{01}$, the
      weight ratio of '0' and '1' tokens. The y axis is $w_{02}$ the
      weight ratio of '0' and '2' tokens. The plots show results for
      three different random initializations.}
    \label{fig:head_performance_x}
  \end{center}
\end{figure}

\section{Head Performance at Initialisation and During Training} 
\label{app:sacc_evolution}

\begin{figure}[htb]
  \begin{center}
    \includegraphics[width=0.48\linewidth]{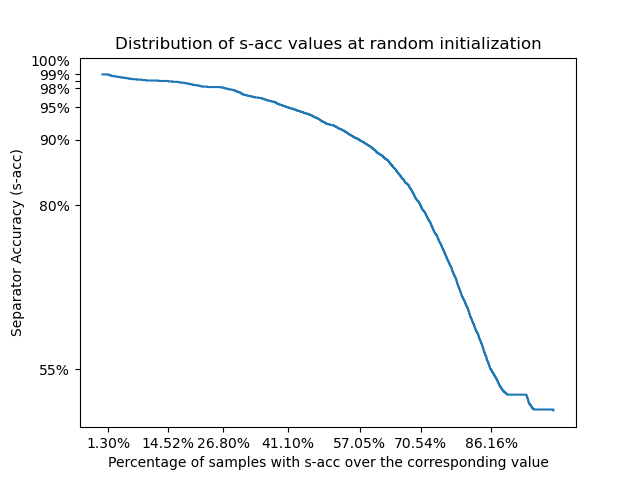}
    \includegraphics[width=0.48\linewidth]{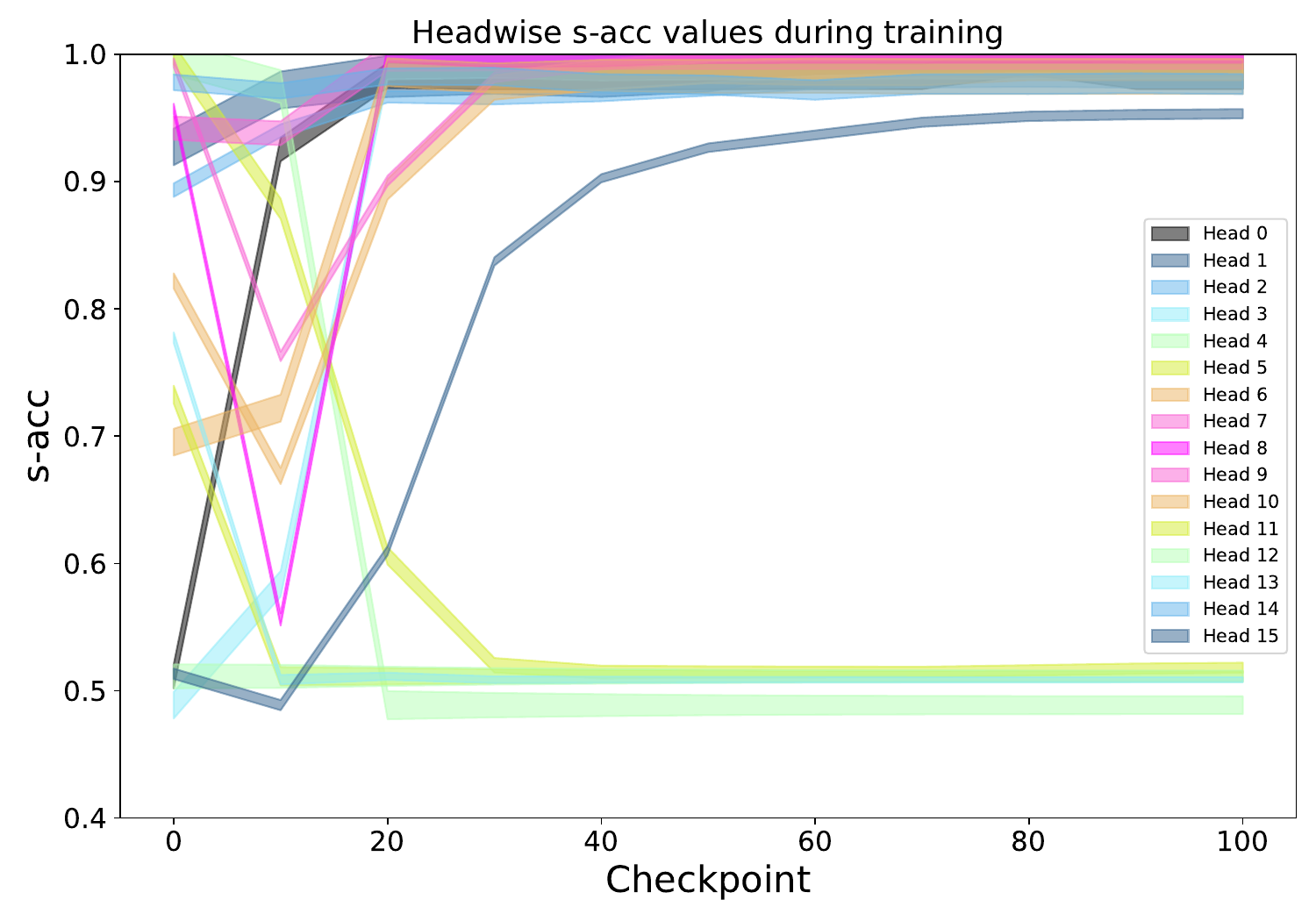}
    \caption{\textbf{Left:} Distribution of head s-acc values without training, after random initialization, based on 10000 heads. \textbf{Right:}
      Headwise s-acc values during training. Line width indicates the weight with which the output layer attends to the given head.}
    \label{fig:sacc_evolution}
  \end{center}
\end{figure}

Figure~\ref{fig:sacc_evolution} shows the distribution of s-acc values
in randomly initialised heads, as well as their evolution during
learning. We find that even without training, a large number of heads
perform reasonably well, even if perfect separation is only achievable
in $1.3\%$ of the heads. During training, the s-acc distribution
becomes extremely bipolar with very strong and very weak heads.

\section{Outputs for Individual Heads}
\label{app:task1}
\begin{figure}[htb]
  \begin{center}
    \includegraphics[width=0.23\linewidth]{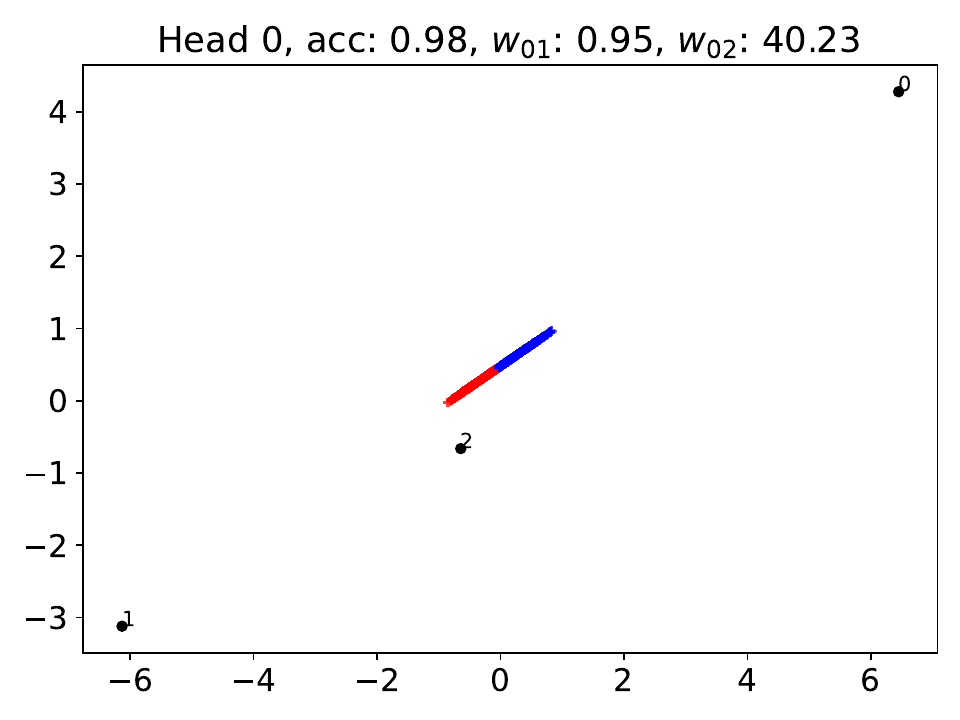}
    \includegraphics[width=0.23\linewidth]{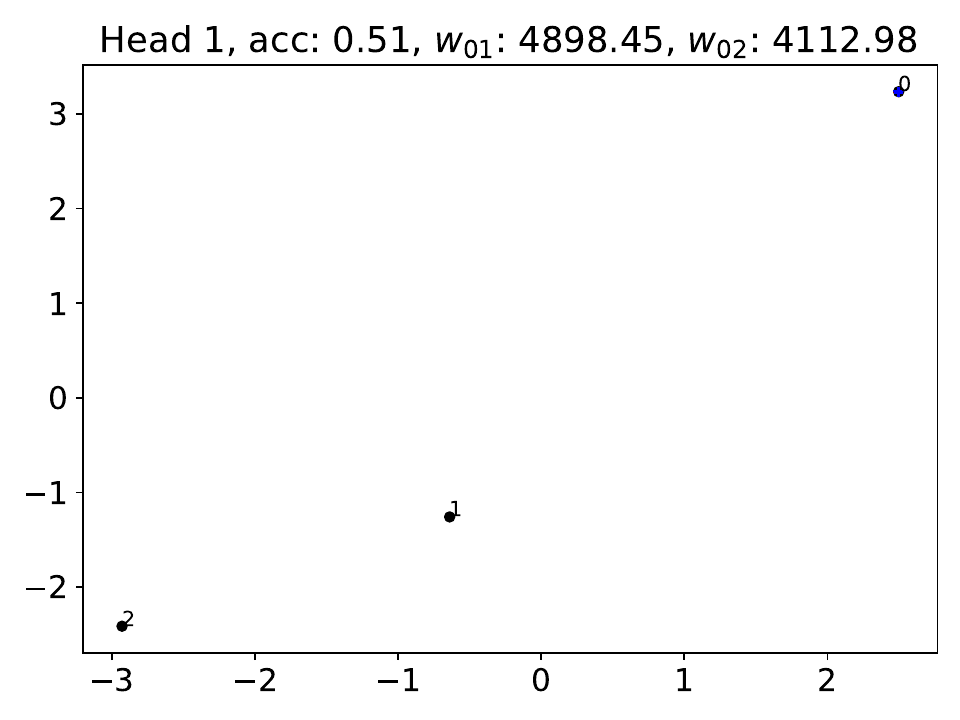}
    \includegraphics[width=0.23\linewidth]{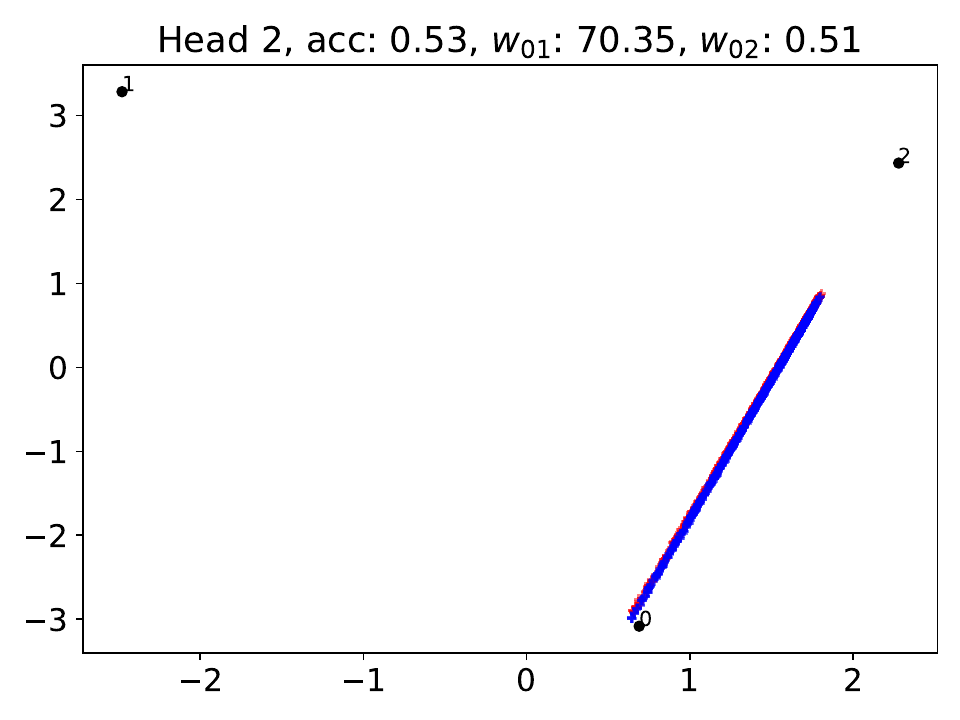}
    \includegraphics[width=0.23\linewidth]{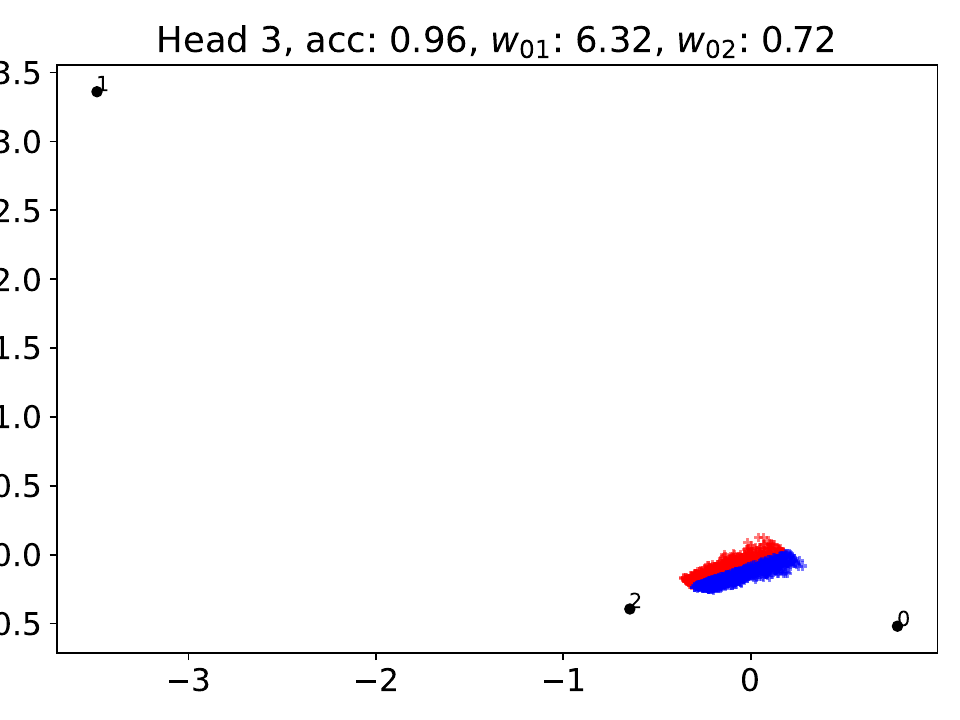}
    \includegraphics[width=0.23\linewidth]{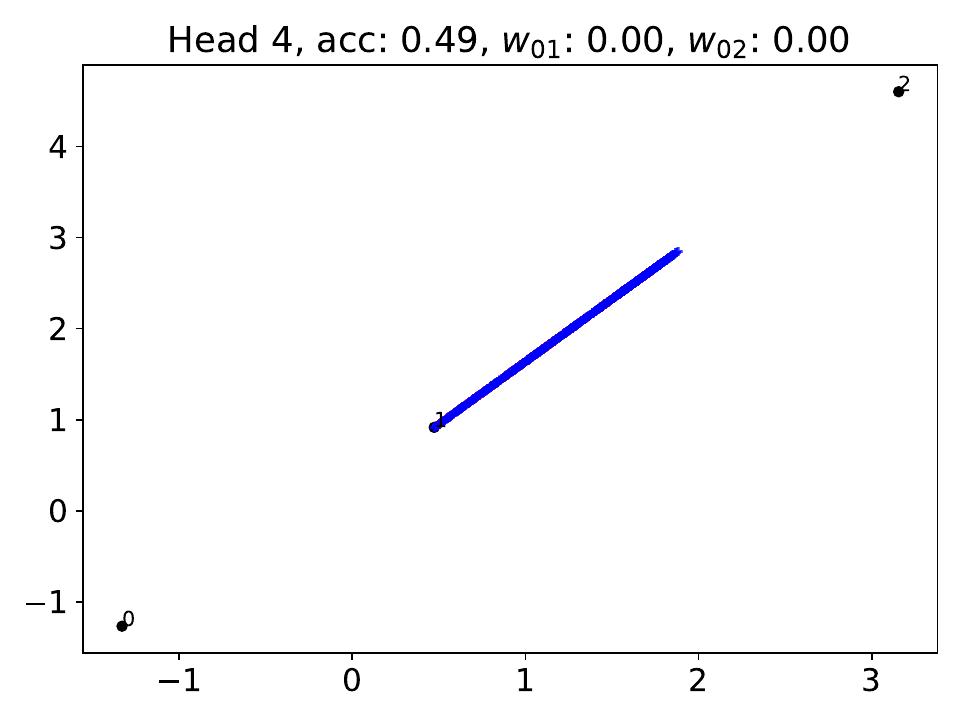}
    \includegraphics[width=0.23\linewidth]{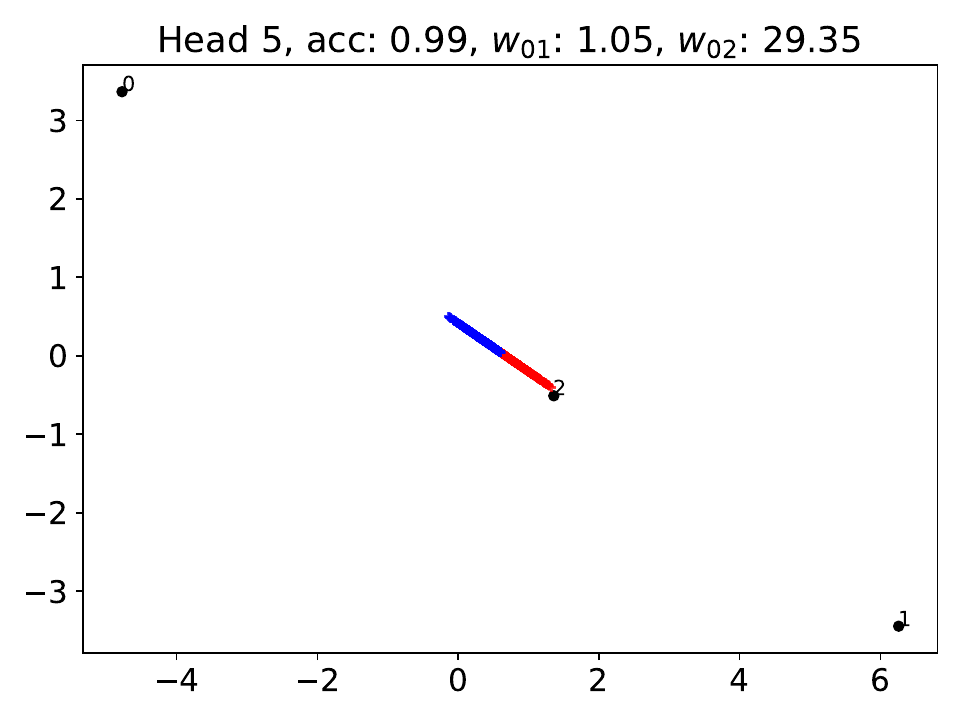}
    \includegraphics[width=0.23\linewidth]{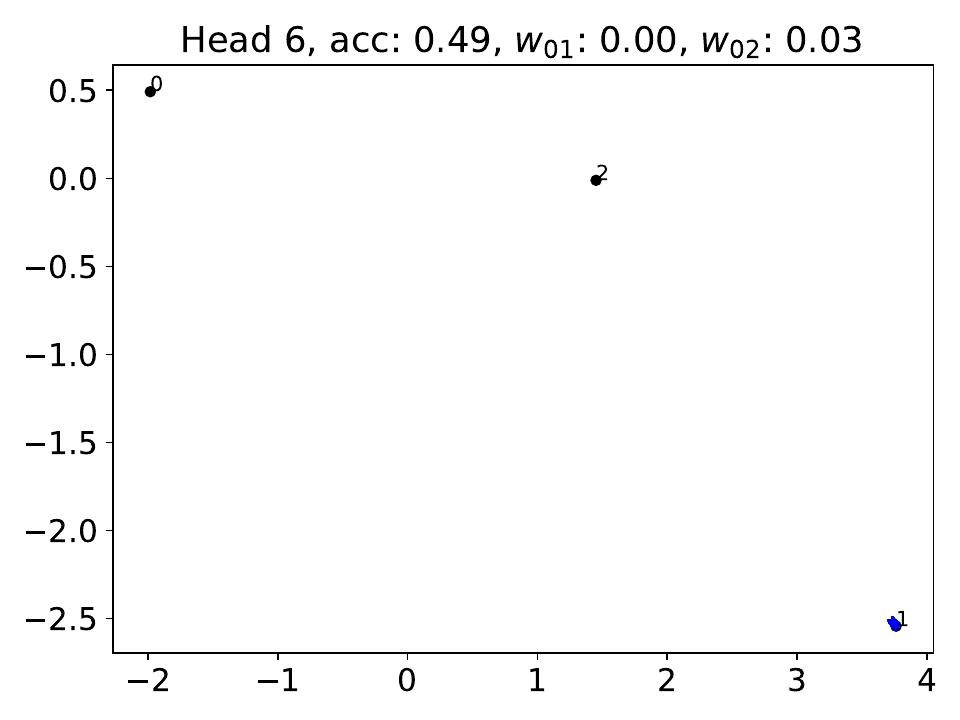}
    \includegraphics[width=0.23\linewidth]{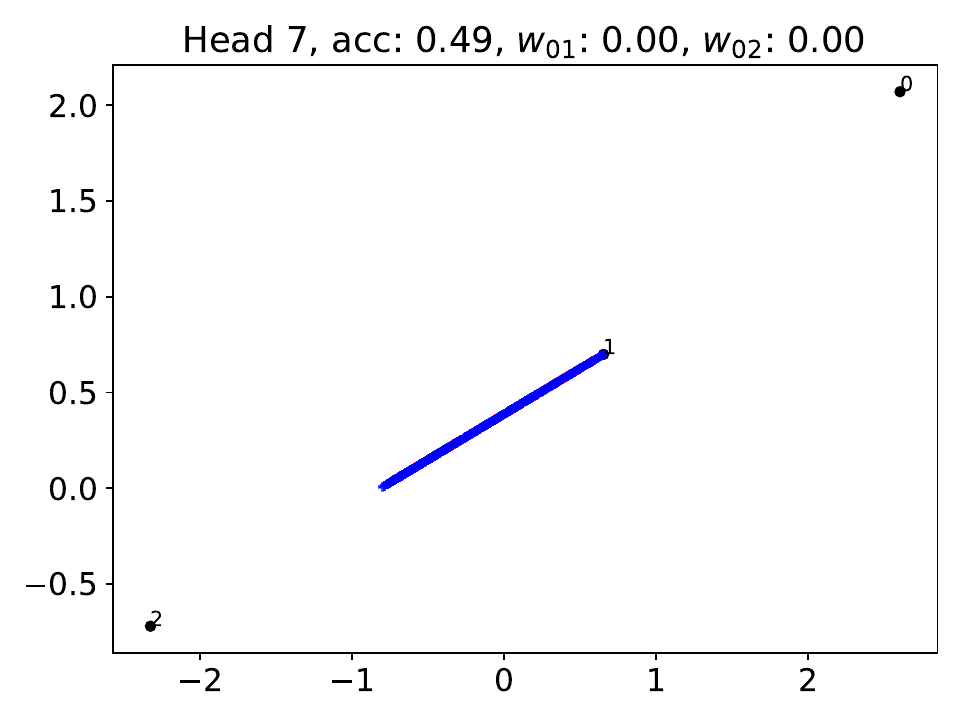}
    \includegraphics[width=0.23\linewidth]{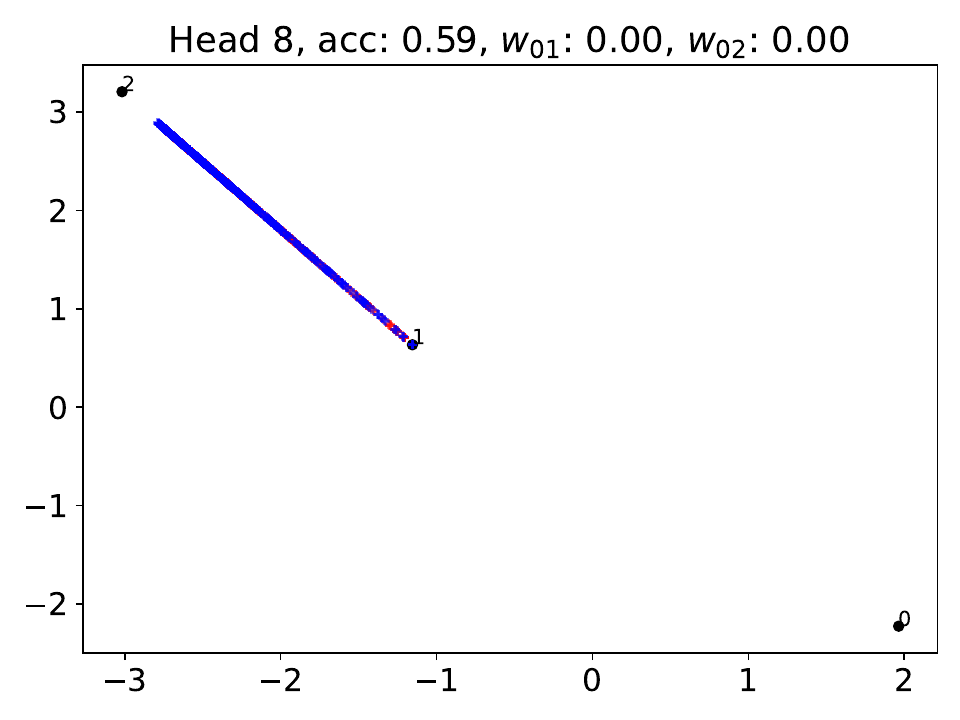}
    \includegraphics[width=0.23\linewidth]{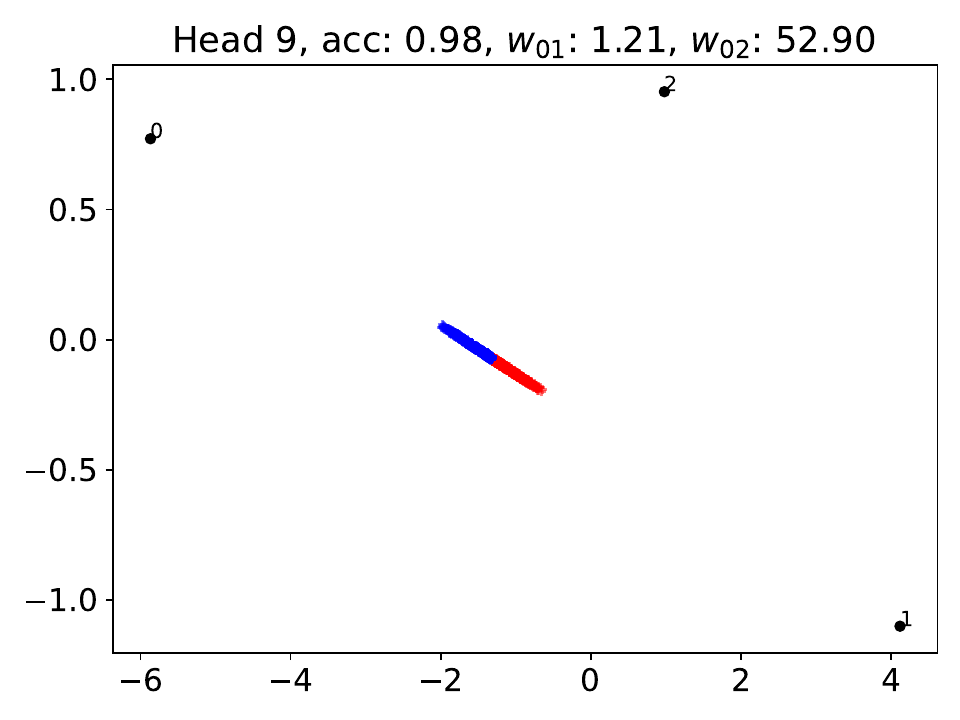}
    \includegraphics[width=0.23\linewidth]{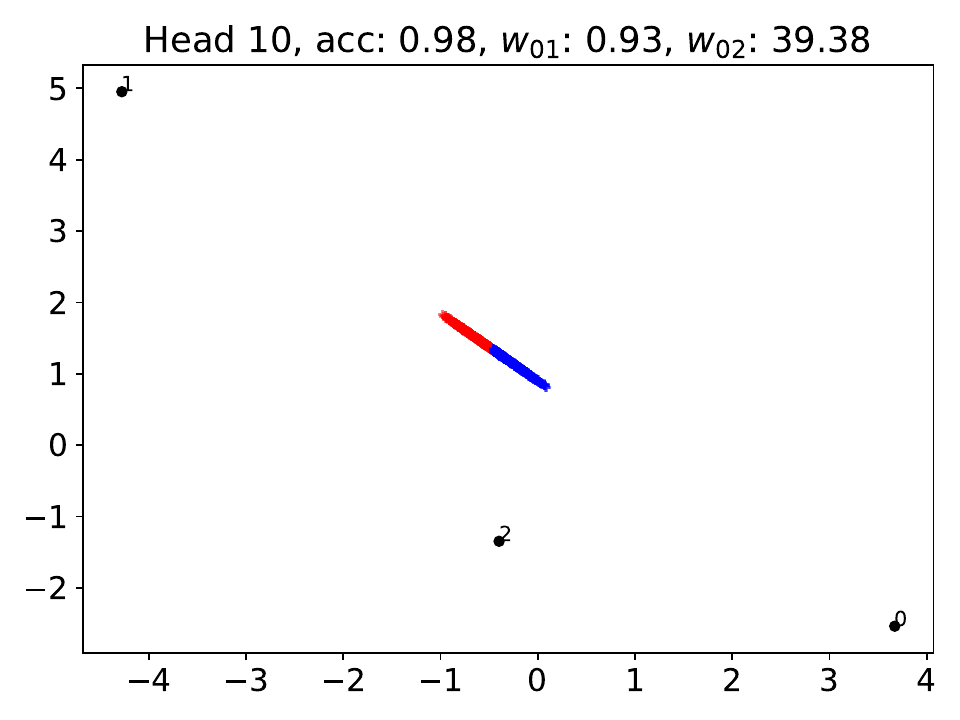}
    \includegraphics[width=0.23\linewidth]{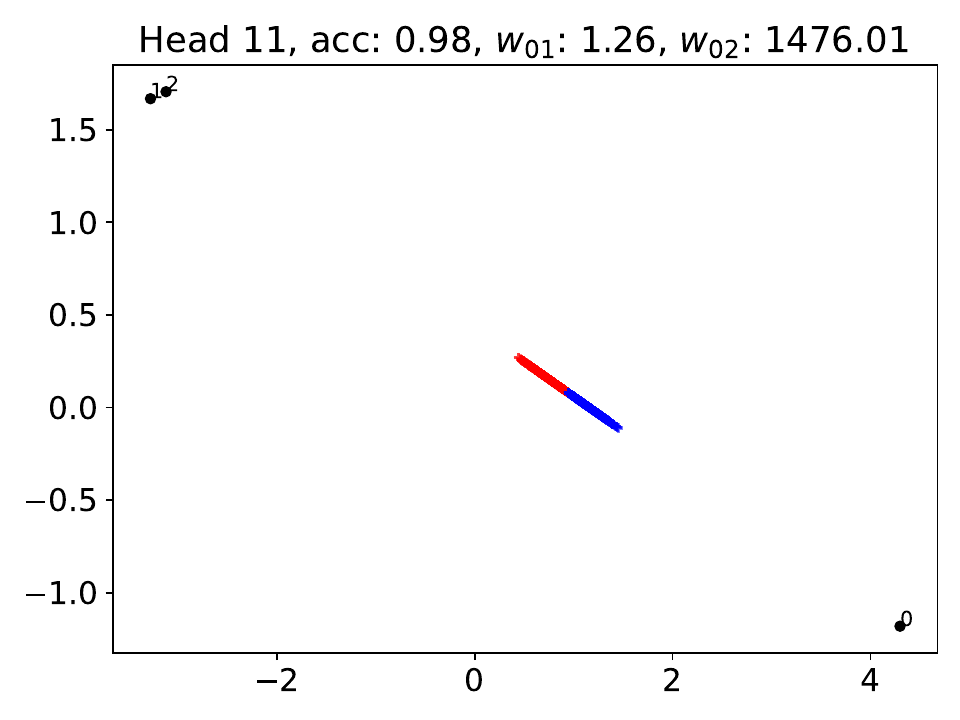}
    \includegraphics[width=0.23\linewidth]{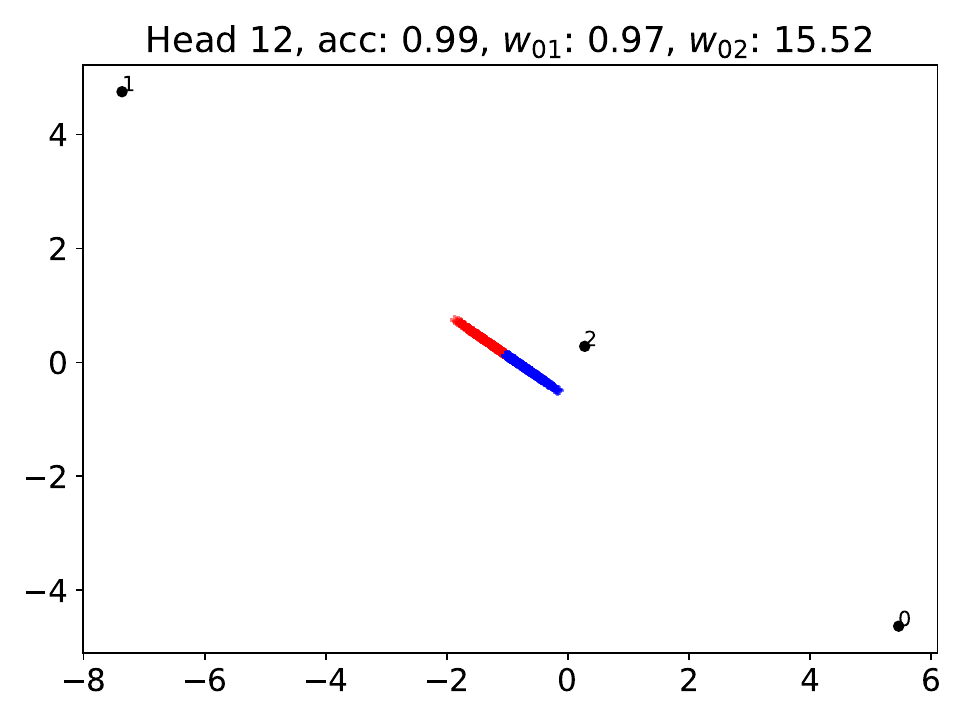}
    \includegraphics[width=0.23\linewidth]{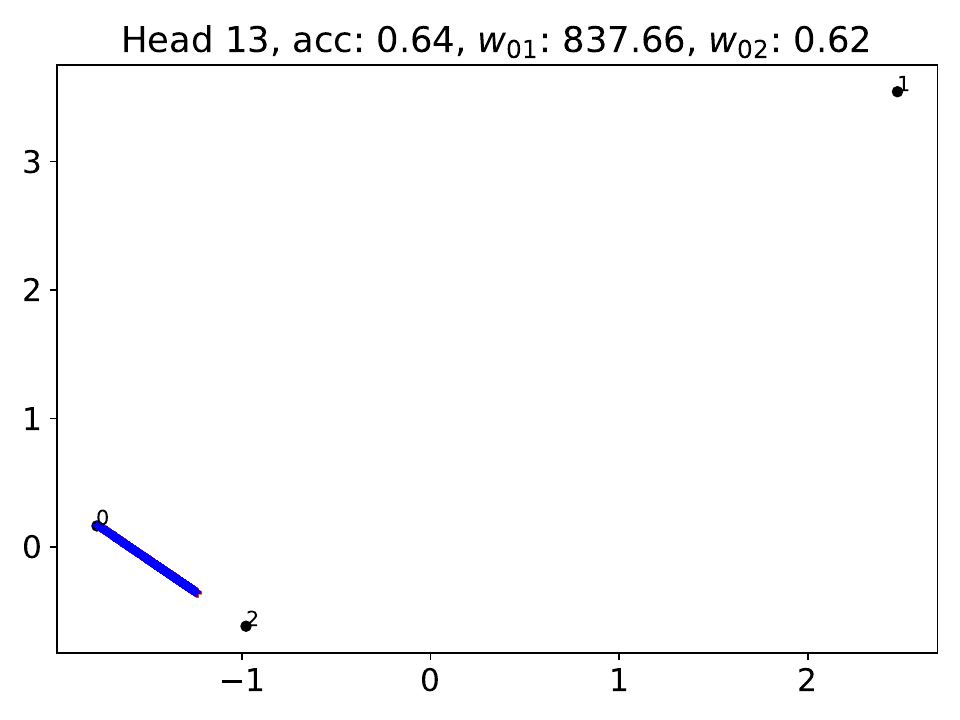}
    \includegraphics[width=0.23\linewidth]{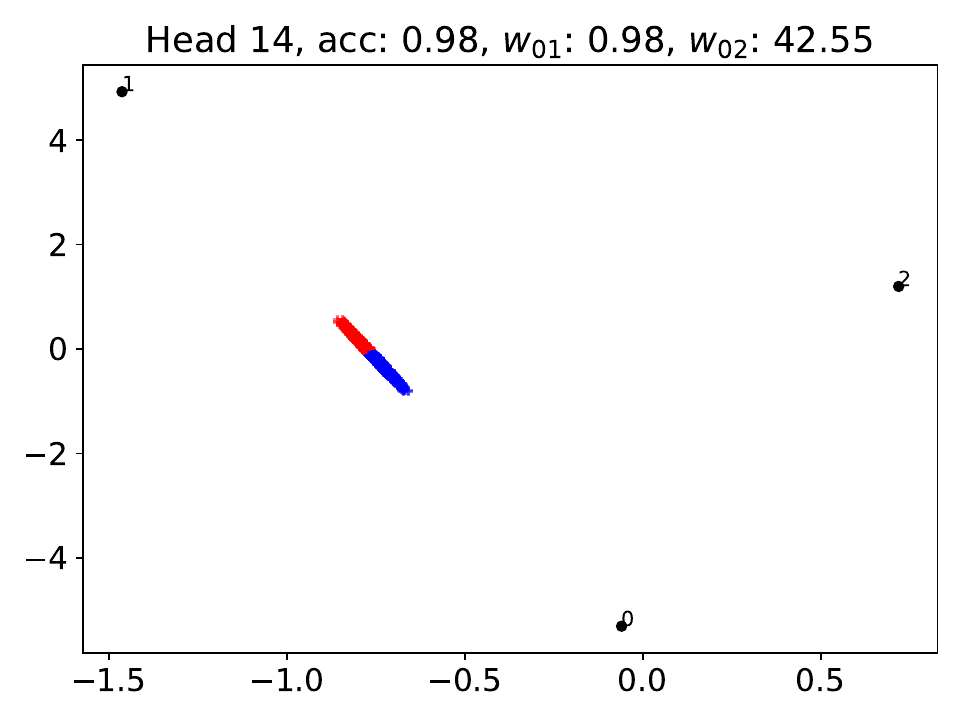}
    \includegraphics[width=0.23\linewidth]{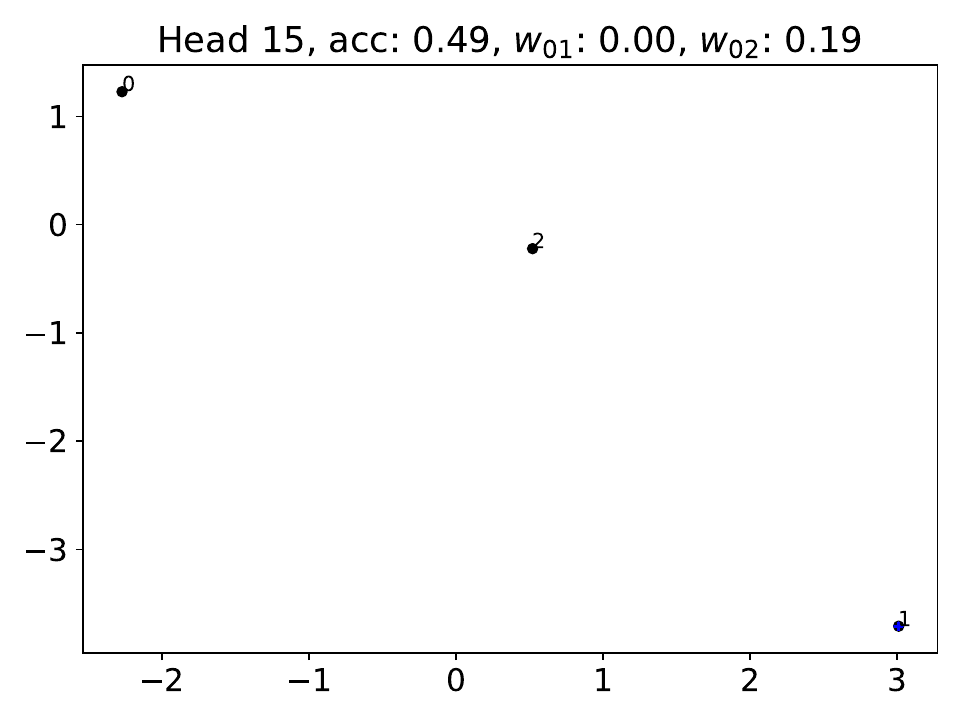}
  \end{center}
  \caption{Colored dots indicate head outputs for the test samples
    with red for samples having more '1's and blue for the rest. We
    also show the value feature vectors of the '0', '1', '2'
    tokens.}  \label{fig:head_outputs}
\end{figure}

In Figure~\ref{fig:head_outputs}, we plot the head outputs for the test samples, using colors blue and red to distinguish classes. We also plot the value feature vectors for '0', '1', '2' tokens.

\end{document}